\documentclass[10pt,twocolumn,letterpaper]{article}

\usepackage{ieee}
\usepackage{times}
\usepackage{epsfig}
\usepackage{graphicx}
\usepackage{amsmath}
\usepackage{amssymb}
\usepackage{mathtools}
\usepackage{subfigure}
\usepackage{booktabs} % for professional tables
\usepackage{algorithm}
\usepackage{algpseudocode}
\usepackage{listings}

\usepackage{etoolbox}
\makeatletter
\AfterEndEnvironment{algorithm}{\let\@algcomment\relax}
\AtEndEnvironment{algorithm}{\kern2pt\hrule\relax\vskip3pt\@algcomment}
\let\@algcomment\relax
\newcommand\algcomment[1]{\def\@algcomment{\footnotesize#1}}
\renewcommand\fs@ruled{\def\@fs@cfont{\bfseries}\let\@fs@capt\floatc@ruled
  \def\@fs@pre{\hrule height.8pt depth0pt \kern2pt}%
  \def\@fs@post{}%
  \def\@fs@mid{\kern2pt\hrule\kern2pt}%
  \let\@fs@iftopcapt\iftrue}
\makeatother
\usepackage[pagebackref=true,breaklinks=true,letterpaper=true,colorlinks,bookmarks=false]{hyperref}

\iccvfinalcopy % *** Uncomment this line for the final submission

\ificcvfinal\fi

\begin{document}

%%%%%%%%% TITLE
\title{F-Drop\&Match: GANs with a Dead Zone in the High-Frequency Domain}

\author{Shin'ya Yamaguchi\\
NTT\\
{\tt\small shinya.yamaguchi.mw@hco.ntt.co.jp}
\and
Sekitoshi Kanai\\
NTT\\
{\tt\small sekitoshi.kanai.fu@hco.ntt.co.jp}
}

\maketitle

%%%%%%%%% ABSTRACT
\begin{abstract}
   Generative adversarial networks built from deep convolutional neural networks (GANs) lack the ability to exactly replicate the high-frequency components of natural images.
   To alleviate this issue, we introduce two novel training techniques called frequency dropping (F-Drop) and frequency matching (F-Match).
   The key idea of F-Drop is to filter out unnecessary high-frequency components from the input images of the discriminators.
   This simple modification prevents the discriminators from being confused by perturbations of the high-frequency components.
   In addition, F-Drop makes the GANs focus on fitting in the low-frequency domain, in which there are the dominant components of natural images.
   F-Match minimizes the difference between real and fake images in the frequency domain for generating more realistic images.
   F-Match is implemented as a regularization term in the objective functions of the generators; it penalizes the batch mean error in the frequency domain.
   F-Match helps the generators to fit in the high-frequency domain filtered out by F-Drop to the real image.
   We experimentally demonstrate that the combination of F-Drop and F-Match improves the generative performance of GANs in both the frequency and spatial domain on multiple image benchmarks.
\end{abstract}

\section{Introduction}
\label{sec:intro}

\begin{figure}[t]
   \centering
   \includegraphics[width=1.0\columnwidth]{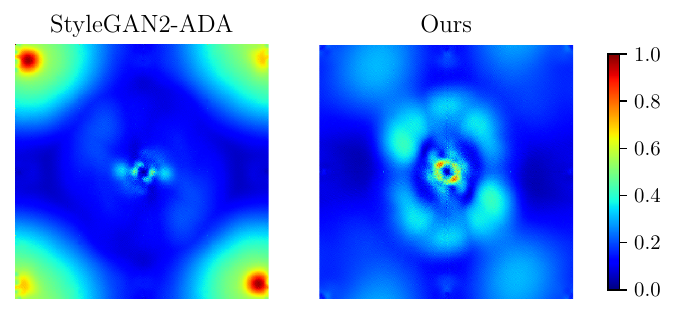}
   \caption{
      Sensitivities of discriminators in the frequency domain.
      The sensitivity is measured by single Fourier attack (SFA)~\cite{tsuzuku_CVPR2019_structural_sensitivity}, which perturbs each frequency component of an image.
      As the sensitivity, we plot the average differences between outputs of normal and attacked discriminators over 128 images in the AFHQ-Cat dataset (\(512\times 512\)) on each pixel.
      The differences in the low-frequency domain are located near the center of each figure, and the differences in the high-frequency domains are at the edges.
      Our method outperforms the baseline (StyleGAN2-ADA) in terms of the robustness against the SFA on the high-frequency components.
   }
   \label{fig:sfa_comp}
 \end{figure}

Generative adversarial networks built from deep convolutional networks~(GANs)~\cite{Goodfellow_NIPS14,gulrajani17_wgangp,miyato_SNGAN_iclr18,Radford_ICLR16_DCGAN} have attracted much attention in the computer vision community and have been utilized in various applications because they can synthesize diverse images with high-fidelity to the target datasets.
The training of GANs is formulated as a competitive game played by two neural networks called a generator and a discriminator; the generator is optimized to produce fake images that can fool the discriminator, and the discriminator is optimized to distinguish the real images from the fake images through min-max optimization.
In theory, the model replicates training data as the optimal result.
However, recent studies have revealed that GANs fail to replicate data in the frequency domain~\cite{durall_CVPR20_watch_your_upconv,frankICML20_leveraging_frequency}.
Durall~\etal~\cite{durall_CVPR20_watch_your_upconv} and Frank~\etal~\cite{frankICML20_leveraging_frequency} have reported that the frequency characteristics of the generated images in the high-frequency domain are different from those of real images (we refer to this difference as the frequency gap).
They have also shown that the generated images can be easily detected as fakes with almost 100\% accuracy by assessing the frequency gap.
While the previous studies mainly focus on the aliasing caused by upsampling in CNNs as the cause of the frequency gap, modifying the upsampling is insufficient for correcting the flaws in the frequency domain~\cite{frankICML20_leveraging_frequency}.
In this study, we explore another cause of the frequency gaps to reduce them.
Since spatial and frequency domains are dual, reducing the frequency gaps can improve the generative performances of GANs in the spatial domain.

We hypothesize that the frequency gap is caused by the sensitivity of the discriminators to the perturbations in the high-frequency domain.
In GANs for image generation, discriminators are usually implemented as CNN-based binary classifiers.
As shown in~\cite{tsuzuku_CVPR2019_structural_sensitivity,yin_NIPS2019_fourier}, CNN-based classifiers are sensitive to perturbations of the frequency components.
Moreover, Wang~\etal~\cite{wang_CVPR2020_high_frequency} have reported that CNN-based classifiers predict labels depending on high-frequency components that are hardly recognizable to humans.
Accordingly, we conjecture that the discriminators of GANs are also sensitive to the high-frequency components of the input images.
Indeed, our experiments demonstrate the sensitivity of the discriminator in the frequency domain: the output of the discriminator is significantly changed by single Fourier attack~\cite{tsuzuku_CVPR2019_structural_sensitivity}, which perturbs each frequency component of an image (Fig.~\ref{fig:sfa_comp}, left).
The sensitivity of the discriminators prevents the generators from learning data because the generators are optimized to fool the discriminators by perturbing high-frequency components rather than by replicating data.

To alleviate the sensitivity of the discriminators and the frequency gap, we present two novel techniques, called {\em frequency dropping} (F-Drop) and {\em frequency matching} (F-Match).
The main idea of F-Drop is to filter out high-frequency components from the inputs of the discriminators (for both real and generated images) and thereby the discriminators concentrate on lower frequency components, which are the dominant components in natural images~\cite{xu_CVPR2020_learning_in_frequency}.
We insert a low-pass filter, which filters out frequency components above a certain threshold from images, before the input layer of the discriminators.
F-Drop, (i) transforms RGB images into the frequency domain by using discrete cosine transform (DCT), (ii) performs filtering in the frequency domain by element-wise multiplication, and (iii) transforms the images back into RGB space by using inverse discrete cosine transform (IDCT).
Since RGB images are used as input, F-Drop does not require any modifications to the original network architectures.
By applying F-Drop, the discriminators become robust against high-frequency perturbations (Fig.~\ref{fig:sfa_comp}, right), and thus, the generators can dedicate themselves to fooling the discriminators by learning the remaining lower frequency components.
However, since F-Drop simply transforms the input of the discriminators, the generators are still free to synthesize the high-frequency components filtered out during the training.
Hence, to synthesize realistic frequency components, we propose F-Match, which minimizes the mean error in the frequency domain.
F-Match is a simple mini-batch-based regularization term for the objective function of the generators; it can utilize arbitrary frequency transformations (\eg, DFT and DCT) and loss functions (\eg, the squared and absolute error).
We experimentally found that the best function for F-Match is the mean squared error in DCT space.
Our experiments show that, in various settings, the combination of F-Drop and F-Match succeeds in synthesizing more realistic images in both the frequency and spatial domains compared with the conventional techniques~\cite{chen_AAAI21_SSDGAN,durall_CVPR20_watch_your_upconv,frankICML20_leveraging_frequency}.
Our contributions are summarized as follows:
\begin{itemize}
   \item We demonstrate that the discriminators of GANs are sensitive to perturbations of high-frequency components through the experiments applying single Fourier attack to discriminators.
   \item We propose two simple techniques for GANs called F-Drop and F-Match for reducing the frequency gap between real and generated images. F-Drop filters out the high-frequency components from the input images of the discriminators, and F-Match minimizes the mean error in the frequency domain by adding a regularization term to the objective function of the generators.
   \item We confirm that our methods can improve the quality of the generated images on various image datasets.
\end{itemize}

\section{Related Work}
\label{sec:related}
\subsection{Frequency Gaps in Generative Models}\label{sec:related_freq_gap}
Frequency gaps in GANs or CNN-based generative models have been studied in recent papers~\cite{durall_CVPR20_watch_your_upconv,frankICML20_leveraging_frequency}.
Durall~\etal~\cite{durall_CVPR20_watch_your_upconv} and Frank~\etal~\cite{frankICML20_leveraging_frequency} have found that there are frequency gaps between real images and images generated from CNN-based models by using discrete Fourier transform (DFT) and discrete cosine transform (DCT).
They have also found that the generated images are detected as fake by linear classifiers trained on the frequency components of the images.
These studies have hypothesized that upsampling in CNNs is a cause of the frequency gaps.
In particular, Frank~\etal~\cite{frankICML20_leveraging_frequency} have shown that the frequency gaps can be reduced by modifying the upsampling in the generators (by using, \eg, binomial upsampling).
However, the modification of upsampling is not sufficient for generating undetectable fake images in the frequency domain.
Furthermore, we empirically report that the modification may degrade the generative performance of GANs in the spatial domain (Sec.~H of the supplementary materials); this degradation has not been discussed in any previous works.
In contrast to these works, we show that the discriminators of GANs are sensitive to high-frequency perturbations, and that this sensitivity is also one of the causes of the frequency gaps.

For alleviating frequency gaps, Durall~\etal~\cite{durall_CVPR20_watch_your_upconv} have proposed spectral regularization which minimizes the binary cross-entropy between the azimuthal integrals of the real and generated images in the frequency domain.
Although spectral regularization has a similar form to F-Match, it minimizes the gaps between each generated image and the mean value of the real images whereas F-Match minimizes the gaps between the mean values of the generated and real images over each mini-batch.
Chen~\etal~\cite{chen_AAAI21_SSDGAN} have proposed a similar technique called SSD, which modifies discriminators by adding a classifier in the frequency domain and utilizes the output of the classifier to modulate the losses of the GANs.
SSD does not use the gradients of the frequency classifier for training of GANs, whereas F-Match directly uses the gradients of the loss in the frequency domain.

\subsection{Sensitivity of CNNs for Frequency Components}\label{sec:sensitivity_cnn}
In the context of adversarial attacks for CNN models, Tsuzuku and Sato~\cite{tsuzuku_CVPR2019_structural_sensitivity} have pointed out a sensitivity of CNNs in the frequency domain by conducting an analysis using their own black-box attack called single Fourier attack (SFA).
SFA perturbs an image in the directions of each Fourier basis.
Similar to \cite{tsuzuku_CVPR2019_structural_sensitivity}, Yin~\etal~\cite{yin_NIPS2019_fourier} have shown that naturally trained CNNs are sensitive to high-frequency perturbations.
In addition, Wang~\etal~\cite{wang_CVPR2020_high_frequency} have indicated that the output of CNN-based classifiers depends on the high-frequency components that are not visible to humans.
However, they have also shown that dropping the high-frequency components from the training does not degrade the final test performances.
Moreover, Xu~\etal~\cite{xu_CVPR2020_learning_in_frequency} have shown that dropping the high-frequency components from the input images of CNNs by thresholds helps to reduce the input size and improves performance.
In summary, the previous results provide two key insights: (i) CNNs-based classifiers have flaws in processing high-frequency components in input images, and (ii) high-frequency components are not essential for training the classifiers.
These insights underlie the idea of F-Drop described in Sec.~\ref{sec:idea_drop}.

\section{Background}
\label{sec:background}
\subsection{Generative Adversarial Networks}
A generative adversarial network is composed of a {\it generator} network $G_{\theta}$ parametrized by \(\theta\), and a {\it discriminator} network $D_{\phi}$ parameterized by \(\phi\)~\cite{Goodfellow_NIPS14}.
The $G_{\theta}$ generates a fake sample \(x_{\text{fake}}=G_{\theta}(z)\) from a random noise $z\sim p_z$, and the $D_{\phi}$ distinguishes an observation $x$ whether $x$ comes from the data distribution $p_{\rm data}$ or not.
The objective functions for training the discriminator and generator are
\begin{eqnarray}
    \label{eq:adv_loss_d}
    \mathcal{L}_{D_{\phi}} &=& - \mathbb{E}_{x \sim p_{\rm data}}\log{D_{\phi}(x)}    \nonumber \\
          & & - \mathbb{E}_{z \sim p_z}\log{(1-D_{\phi}(G_{\theta}(z)))}, \\
    \label{eq:adv_loss_g}
    \mathcal{L}_{G_{\theta}} &=& - \mathbb{E}_{z \sim p_z}\log{D_{\phi}(G_{\theta}(z)).} 
\end{eqnarray}
By training of $G_{\theta}$ and $D_{\phi}$, $D_{\phi}$ learns to maximize the probability of assigning a ``real" label to real examples and a ``fake'' label to fake examples, whereas $G_{\theta}$ learns to maximize the probability of $D_{\phi}$'s failure of distinction.
In theory, when $G_{\theta}$ and $D_{\phi}$ converge to the optimal point, the generator network $G_{\theta}$ implicitly replicates $p_{\rm data}$.

In this paper, we mainly focus on GANs built from CNNs.
There are several variants, such as DCGAN~\cite{Radford_ICLR16_DCGAN}, WGAN-GP~\cite{gulrajani17_wgangp}, and SNGAN~\cite{miyato_SNGAN_iclr18}.
We can apply F-Drop and F-Match to any of these variants because they are designed as an additional masking layer in discriminators or as an additional regularization term.

\subsection{Frequency Transformations}\label{sec:freq_trans}
Here, we briefly summarize the foundations of discrete cosine transform (DCT) that is used in F-Drop and F-Match.
Note that, for simplicity, our discussion regards transformations of a gray-scale square image \(X\in \mathbb{R}^{H\times H}\) but it can be easily extended to color images by performing the same computations on each channel.

% Two-dimensional discrete Fourier transform (DFT) for an image \(X \in \mathbb{R}^{H\times H}\) in the spatial domain is defined as:
% \begin{equation}
%    F(u, v)\!=\!\sum_{i=0}^{H-1} \sum_{j=0}^{H-1} X(i, j) \exp\!\left[ -2 \pi {\rm j} \left( \frac{u i}{H}\!+\!\frac{v j}{H} \right)\!\right],
% \end{equation}
% where \((i,j)\) represents a spatial pixel coordinate, \((u,v)\) is a frequency coordinate, and \({\rm j}\) is an imaginary unit.
% DFT produces high-frequency distortions because of the discontinuous boundaries; this is known as {\em end effects} of DFT~\cite{tribolet_IEEE1979_frequency_end_effect}.
% For avoiding the end effects, discrete cosine transform (DCT) are often used since DCT does not have the discontinuous boundaries~\cite{tribolet_IEEE1979_frequency_end_effect}.

Two-dimensional DCT~\cite{ahmed_IEEE1974_dct,gonzalez_book_digital_image_processing} is formulated as follows:
\begin{equation}\label{eq:dct}
      C(u,v) = \frac{2\alpha(u)\alpha(v)}{H}\sum^{H-1}_{i=0}\!\sum^{H-1}_{j=0}\!X(i,j) c(i,j,u,v),
\end{equation}
where \(\alpha(0) = 1/\sqrt{2}, \alpha(t) = 1\) (for \(t \neq 0\)), and
\[
      c(i,j, u,v)=\cos\!\left[\frac{(2i\!+\!1)u\pi}{2H}\right]\!\cos\left[\frac{(2j\!+\!1)v\pi}{2H}\right]\!.\!
\]
where \((i,j)\) represents a spatial pixel coordinate, \((u,v)\) is a frequency coordinate.
This form is called DCT-II.
We choose DCT for F-Drop and F-Match as the default because it does not have discontinuous boundaries that produces high-frequency noise, in contrast to DFT~\cite{tribolet_IEEE1979_frequency_end_effect}.
As the transformation from the frequency domain back to the spatial domain, we use two-dimensional inverse discrete cosine transform (IDCT):
\begin{equation}
   X(i,j) = \frac{2}{H}\sum^{H-1}_{u=0}\!\sum^{H-1}_{v=0}\!\alpha(u)\alpha(v)C(u,v) c(i,j,u,v),
\end{equation}
where \(\alpha(\cdot)\) and \(c(\cdot)\) as the same as in Eq.~(\ref{eq:dct}).

\section{Analyzing GANs with Single Fourier Attack}\label{sec:sfa_analysis}
We hypothesize that the frequency gap of GANs is caused by the sensitivity of the discriminators to perturbations in the high-frequency domain.
To confirm this hypothesis, we analyze GANs subjected to single Fourier attack (SFA)~\cite{tsuzuku_CVPR2019_structural_sensitivity}.
SFA attacks classification models by perturbing the input images in the Fourier basis directions.
For each perturbation, SFA selects a single frequency component and creates striped noise according to the selected component.
A perturbation \(\delta(u,v)\) of the frequency coordinate \((u,v)\) for an \(H\times H\) image is defined as follows:
\begin{equation}\label{eq:sfa}
   \begin{aligned}
      \delta(u,v) =& \epsilon((1+{\rm j})(F_{H})_{u} \otimes (F_{H})_{v} \\
              & + (1-{\rm j})(F_{H})_{H-u} \otimes (F_{H})_{H-v}),
   \end{aligned}
\end{equation}
where \(\epsilon\) is a hyperparameter determining the size of the perturbation, \(F_{H}\) is the matrix of the Fourier basis and \((F_{H})_{i}\) represents the \(i\)-th row of \(F_H\). Note that \(\otimes\) means the Kronecker product and \({\rm j}\) is the imaginary unit.
We will use SFA to investigate the sensitivity of discriminators in the frequency domain.

As a preliminary experiment, we tested ResNet-based SNGANs~\cite{miyato_SNGAN_iclr18} trained on the CelebA dataset~\cite{liu_ICCV15_celeba} (The details of the training are shown in Sec.~\ref{sec:setup}).
For SFA, we set \(\epsilon\) to \(10/255\).
The left-hand side of Figure~\ref{fig:sfa_comp} visualizes the results of SFA.
The visualization procedure followed~\cite{tsuzuku_CVPR2019_structural_sensitivity}.
Each pixel coordinate corresponds to a frequency component used for SFA, and each pixel represents the absolute differences \(|D(x)-D(x+\delta(u,v))|\), \ie, the sensitivity to the perturbation.
Note that the differences are normalized to \([0,1]\) by dividing with the maximum value of the SNGANs and ours.
We can see that the SNGANs are sensitive to high frequency perturbations like the results in~\cite{yin_NIPS2019_fourier}.
This indicates that the discriminators are easily fooled by perturbing the high-frequency domain and their sensitivity in this regard leads to the frequency gaps because the generators focus on synthesizing the high-frequency perturbations rather than a realistic image.
Since the high-frequency components are not necessary for training CNNs, as discussed in Sec.~\ref{sec:sensitivity_cnn}, we will examine a method of filtering out them from the input images.

\begin{figure}[t]
   \centering
   \includegraphics[width=1.0\columnwidth]{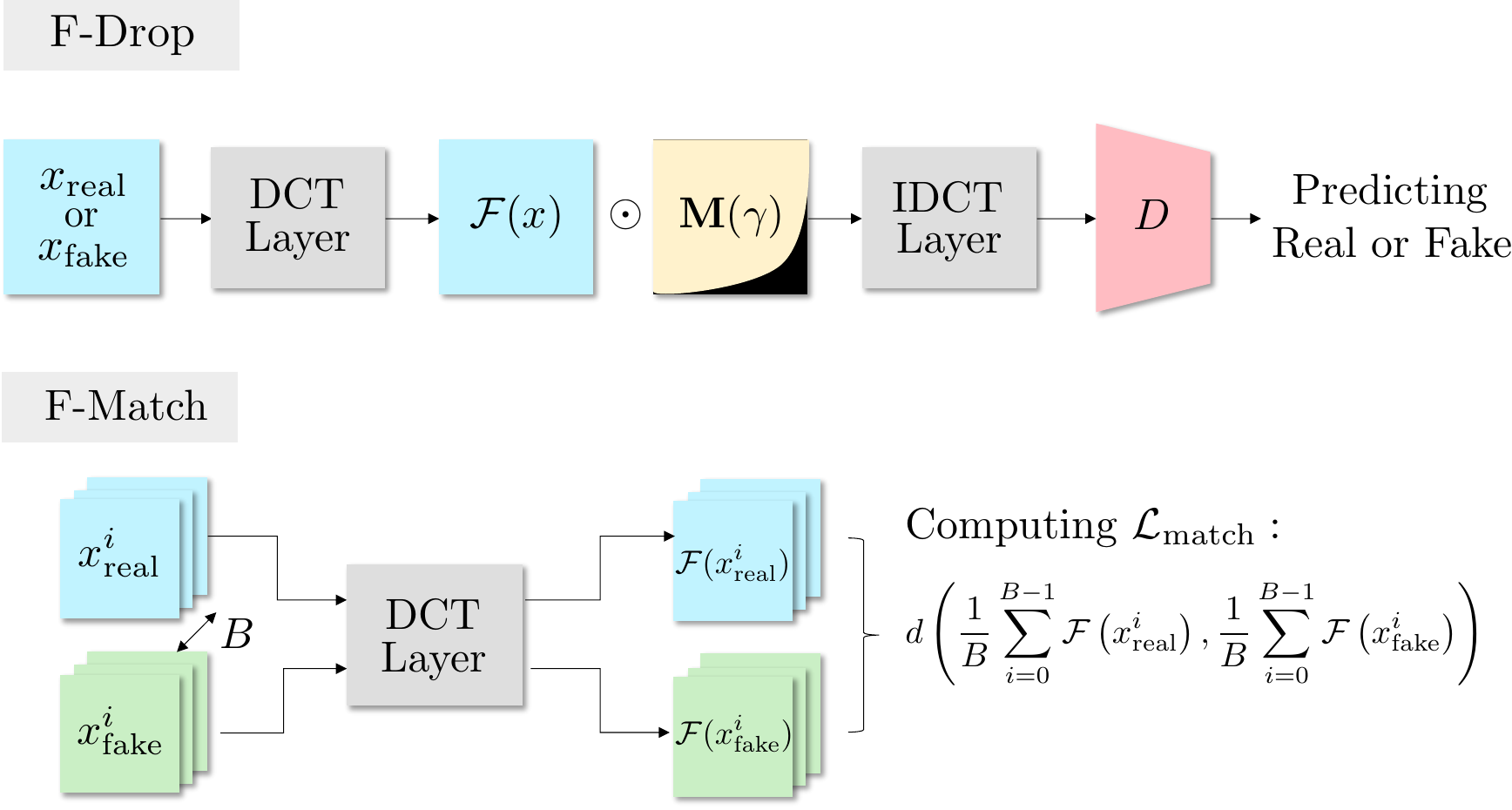}
   \caption{
    Illustration of proposed methods
   }
   \label{fig:proposed_overview}
 \end{figure}

 \section{Proposed Method}
\label{sec:proposed}
Figure~\ref{fig:proposed_overview} illustrates the overview of F-Drop and F-Match.
F-Drop filters out the high-frequency components from the input images for discriminators, while F-Match is a regularization method for generators, which penalizes the mini-batch mean error in the frequency domain between the real and generated images.
F-Drop and F-Match are independent of each other and can be easily incorporated in the architectures of GANs.

\subsection{Frequency Dropping}
\label{sec:idea_drop}
First, we introduce the idea of frequency dropping (F-Drop).
As discussed in Sec.~\ref{sec:sensitivity_cnn} and~\ref{sec:sfa_analysis}, the discriminators of GANs are sensitive to the high-frequency components of the input images, but can be trained without the high-frequency components.
F-Drop is based on these insights; it filters out the high-frequency components from images by masking with a user-defined threshold parameter \(\gamma\in [0,1]\).
The procedure of F-Drop is quite simple: (i) transform an input image into the frequency domain, (ii) drop the high-frequency components, and (iii) transform the frequency components back into the spatial domain.
F-Drop transforms an input color image \(\mathbb{R}^{3\times H \times W}\) for the discriminators as follows:
\begin{equation}
   \operatorname{Drop}(x,\gamma) = \mathcal{F}^{-1}\left(\mathcal{F}(x) \odot \mathbf{M}(\gamma) \right),
\end{equation}
where \(\mathcal{F}\) is a frequency transform function, such as DCT, \(\mathcal{F}^{-1}\) is an inverse frequency transform function, such as IDCT, and \(\mathbf{M}(\gamma)\in\mathbb{R}^{3\times H \times W}\) is a mask matrix for filtering specific frequency components.
Note that \(\odot\) denotes element-wise multiplication.
We chose DCT to be \(\mathcal{F}\) and IDCT to be \(\mathcal{F}^{-1}\).
An element in a coordinate \((c,u,v)\) of the mask matrix \(\mathbf{M}(\gamma)\) is defined as
\begin{equation}\label{eq:mask_element}
   M^c_{u,v}(\gamma) = 
   \begin{cases}
      1 & (\sqrt{u^2 + v^2} \leq \gamma\sqrt{H^2 + W^2}) \\
      0 & (\sqrt{u^2 + v^2} > \gamma\sqrt{H^2 + W^2}).
   \end{cases}
\end{equation}
That is, we drop the high-frequency components of coordinates farther away than \(\gamma\sqrt{H^2 + W^2}\), which is the Euclidean distance from the origin point \((0,0)\) (\ie, the direct current component).
After masking, we utilize the remaining lower frequency components for training the GANs (Fig.~\ref{fig:proposed_overview}, top).
We can adjust the cutoff frequency with the threshold hyperparameter \(\gamma\).
As defined in Eq.~(\ref{eq:mask_element}), the channels \(c\) share the mask element, and thus, the \(\operatorname{Drop}(\cdot)\) calculation can be implemented by broadcasting a single channel mask \(\mathbf{M}(\gamma)\in \mathbb{R}^{H\times W}\).
Since all of the operations in \(\operatorname{Drop}(\cdot)\) are differentiable with respect to the input data, we can train the models by gradient descent via backpropagation in an end-to-end fashion.

\subsection{Frequency Matching}\label{sec:freq_match}
Frequency matching (F-Match) is for minimizing the frequency gap between the real and generated images.
The key idea is matching the frequency characteristics of the real and generated images.
F-Match minimizes the frequency gaps by using the mini-batch statistics of the images because an image generated from GANs does not have a one-to-one correspondence to a real image.
This regularization strategy is commonly used by methods such as feature matching in~\cite{Salimans_NIPS16} and MMD-GAN~\cite{Li_NIPS2017_MMDGAN}.
The loss function of F-Match is formalized as follows:
\begin{equation}\label{eq:fmatch}
      \mathcal{L}_{\text{match}} = d\left(\bar{X}_{\text{real}}, \bar{X}_{\text{fake}} \right),
\end{equation}
\[
   \bar{X}_{\text{real}} = \frac{1}{B} \sum^{B-1}_{i=0} \mathcal{F} (x^i_{\text{real}}),~
   \bar{X}_{\text{fake}} = \frac{1}{B} \sum^{B-1}_{i=0} \mathcal{F} (x^i_{\text{fake}}),
\]
where \(d(\cdot)\) is an error function, \(B\) is batch size for each training iteration, and \(x_{\text{real}}^i\) is the \(i\)-th real image, and \(x_{\text{fake}}^i\) is the \(i\)-th generated image from the GANs.
\(d(\cdot)\) and \(\mathcal{F}(\cdot)\) can be set to an arbitrary error function (\eg, squared error) and arbitrary frequency transform (\eg, DCT).
In the supplementary materials, we evaluate various combinations of \(d(\cdot)\) and \(\mathcal{F}(\cdot)\) and show that the mean squared error (MSE) in DCT space is the best choice.
We use the following MSE-based function:
\begin{equation}\label{eq:fmatch_mse}
   d_{\text{MSE}}\!=\!\frac{1}{HW}\sum^H_u\sum^W_v\left(\bar{X}_{\text{real}}(u,v) - \bar{X}_{\text{fake}}(u,v) \right)^2,
\end{equation}
where \(\bar{X}(u,v)\) is the \((u,v)\) coordinate of \(\bar{X}\).
In the optimization, \(\mathcal{L}_{\text{match}}\) is added as a regularization term to the objective function defined in Eq.~(\ref{eq:adv_loss_g}):
\begin{equation}
   \mathcal{L}_{G_\theta} = \mathbb{E}_{z \sim p_z}\log{D_{\phi}(G_{\theta}(z))} + \lambda \mathcal{L}_{\text{match}},
\end{equation}
where \(\lambda\) is a balancing hyperparameter.
As discussed in Sec.~\ref{sec:related_freq_gap}, spectral regularization (SR)~\cite{durall_CVPR20_watch_your_upconv} is defined in a similar form to F-Match.
Following~\cite{durall_CVPR20_watch_your_upconv}, the loss function of SR for \(H\times H\) square images is defined as
\begin{eqnarray}
   \label{eq:sr}
   &\mathcal{L}_{\text{SR}}\!=\!\displaystyle\frac{1}{B} \sum^{B-1}_{i=0} d_{\text{SR}}\left(\bar{X}_{\text{real}}, \mathcal{F}(x^i_{\text{fake}})\right),\\
   &d_{\text{SR}}\!=\!-\displaystyle\frac{1}{H/2-1}\!\!\!\sum^{H/2-1}_{r=0}\!\!\operatorname{BCE}\!\left(\!A(\bar{X}_{\text{real}},r),A(X_{\text{fake}},r)\!\right)\!,\! \nonumber
\end{eqnarray}
where \(\operatorname{BCE}(\cdot)\) is the binary cross entropy function and \(A(X,r)\) is the azimuthal integral \(\frac{1}{2\pi}\int^{2\pi}_{0}|X(r,\theta)|d\theta \), which approximates 2D DFT images into 1D signals with respect to the radial distance \(r\) in polar coordinates \((r,\theta)\).
Note that SR differs from F-Match in that it uses a single generated image for minimizing the frequency gaps.

The final objective functions using F-Drop and F-Match are:
\begin{equation}\label{eq:final_loss_d}
   \begin{aligned}
      \mathcal{L}_{D_{\phi}} =& - \mathbb{E}_{x \sim p_{\rm data}}\log{D_{\phi}(\operatorname{Drop}(x,\gamma))}\\
      &  - \mathbb{E}_{z \sim p_z}\log{(1-D_{\phi}(\operatorname{Drop}(G_{\theta}(z),\gamma)))},
   \end{aligned}
\end{equation}
\begin{equation}\label{eq:final_loss_g}
      \mathcal{L}_{G_{\theta}} = - \mathbb{E}_{z \sim p_z}\log{D_{\phi}(\operatorname{Drop}(G_{\theta}(z),\gamma))\!+\!\lambda \mathcal{L}_{\text{match}}.} 
\end{equation}
The overall training procedure with F-Drop and F-Match is summarized in Algorithm~\ref{alg:overall}.
Note that, unlike the training of normal GANs, we pre-fetch the input real images \(\{x_i\}\) for calculating the loss function of F-Match (Eq.~\ref{eq:fmatch}) on line 4.
{\tt GetSample} and {\tt GenNoise} are functions for fetching batch images and for generating batch noise from a normal distribution.

\begin{algorithm}[t]
   \caption{Training of GAN with F-Drop and F-Match}
   \label{alg:overall}
\begin{algorithmic}[1]
{\footnotesize   
   \Require{Batchsize \(B\), learning rate \(\eta_\theta, \eta_\phi\), number of critics \(K\), hyperparameters \(\gamma,\lambda\)}
   \State{Randomly initialize parameters \(\theta,\phi\)}
   \While{not convergent}
   \For{\(k= 1\) to \(K\)}
      \State{\(\{x^i_{\text{real}}\}^{B-1}_{i=0} \leftarrow\) {\tt GetSample}{(\(B\))}}
      \State{\(\{z^i\}^{B-1}_{i=0} \leftarrow\) {\tt GenNoise}{(\(B\))}}
      \If{\(k=1\)}
         \State{\(\{x^i_{\text{fake}}\}^{B-1}_{i=0} \leftarrow \{G_\theta(z^i)\}^{B-1}_{i=0}\)}
         \State{\(\mathcal{L}_{\text{match}}\leftarrow d\left(\frac{1}{B} \sum^{B-1}_{i=0} \mathcal{F} (x^i_{\text{real}}), \frac{1}{B} \sum^{B-1}_{i=0} \mathcal{F} (x^i_{\text{fake}}) \right)\)}
         \State{\(\mathcal{L}_{G_\theta} \leftarrow -\sum^{B}_{i} \log D_{\phi}(\text{Drop}(x^i_{\text{fake}},\gamma)) + \lambda\mathcal{L}_{\text{match}}\)}
         \State{\(\theta \leftarrow \theta - \eta_\theta\nabla_\theta \mathcal{L}_{G_\theta}\)}
      \EndIf{}
      \State{\(\mathcal{L}_{D_\phi} \leftarrow -\sum^{B-1}_{i=0} \log D_{\phi}(\text{Drop}(x_{\text{real}}^i,\gamma))\)}
      \State{~~~~~~~~~~~~\(-\sum^{B-1}_{i=0} \log(1 - D_{\phi}(\text{Drop}(G_\theta(z^i),\gamma)))\)}
      \State{\(\phi \leftarrow \phi-\eta_\phi\nabla_{\phi} \mathcal{L}_{D_\phi}\)}
      \EndFor{}
   \EndWhile{}
}
\end{algorithmic}
\end{algorithm}

\section{Experiments}
\label{sec:results}
We evaluate our proposed methods (F-Drop and F-Match) by comparing them with naive baselines and the existing methods~\cite{chen_AAAI21_SSDGAN,durall_CVPR20_watch_your_upconv,frankICML20_leveraging_frequency}.
We evaluate our methods in terms of 
(i) quantitative metrics for GANs (main evaluations),
(ii) sensitivity to the high-frequency components (frequency sensitivity analysis), 
(iii) the results in the state-of-the-art settings using StyleGAN2-ADA,
and (iv) the quality of the generated images.
The supplementary materials contain an ablation study on F-Match, an analysis of hyperparameter sensitivity of F-Drop, and fake detection in the frequency domain (fake detection).

\subsection{Setup}\label{sec:setup}
\paragraph{Datasets}
We used the six different image datasets: CIFAR-10 and CIFAR-100 (\(32\!\times\!32\))~\cite{krizhevsky09_cifar10}, TinyImageNet (\(32\!\times\!32\))~\cite{wu2017tiny}, STL-10 (\(48\!\times\!48\))~\cite{coates_JMLR11_stl10}, CelebA (\(128\!\times\!128\))~\cite{liu_ICCV15_celeba}, and ImageNet (\(128\!\times\!128\))~\cite{russakovsky_imagenet}.
These datasets have been used for testing the benchmarks of GANs~\cite{brock2018biggan,gulrajani17_wgangp,Lee_WACV21_InfoMaxGAN,miyato_SNGAN_iclr18,Zhang_ICLR20_CRGAN}.
We applied center cropping and resizing to the images of TinyImageNet, CelebA, and ImageNet before the training.
For training, we normalized images into a range of \([-1,1]\).

\paragraph{GAN Baselines}
As a baseline, we chose spectral normalization GAN (SNGAN) with ResNet-backbone architectures~\cite{miyato_SNGAN_iclr18}.
As additional baselines, we tested Binomial~\cite{frankICML20_leveraging_frequency}, which replaces the bilinear upsampling filters in the generators with the low-pass filters based on a binomial distribution, spectral regularization (SR)~\cite{durall_CVPR20_watch_your_upconv}, which minimizes the gaps of the azimuthal integral in DFT space by using Eq.~(\ref{eq:sr}) (see Sec.~\ref{sec:freq_match}), and SSD-GAN~\cite{chen_AAAI21_SSDGAN}, which adds a frequency classifier in DFT space (using the azimuthal integral) to the discriminator and utilizes the output of the classifier to modulate the loss functions of the GANs.
We used the Binomial-5 kernel, following Frank~\etal~\cite{frankICML20_leveraging_frequency}.
SR was based on one in the author's public code repository and used \(\lambda\) as \(1.0\times 10^{-5}\).\footnote{https://github.com/cc-hpc-itwm/UpConv/}\(^{,}\)\footnote{We used PyTorch to implement the differentiable azimuthal integral because the author's implementation, which uses Numpy, is not differentiable. More detailed discussions appear in the supplementary materials.}
For SSD-GAN, the implementation was composed of the author's code, and we used \(\lambda=0.5\), following~\cite{chen_AAAI21_SSDGAN}.\footnote{https://github.com/cyq373/SSD-GAN}
For the labeled datasets (\ie, CIFAR-10/-100, TinyImageNet, and ImageNet), we used conditional batch normalization for generators and the projection discriminator following~\cite{miyato_cgans_iclr18}.
We evaluate other representative GAN variants, including deep convolutional GAN (DCGAN)~\cite{Radford_ICLR16_DCGAN} and Wasserstein GAN with a gradient penalty (WGAN-GP)~\cite{gulrajani17_wgangp} instead of SNGAN in Sec.~\ref{sec:main_result}.
We implemented the architectures of GANs with the open-source repository of Lee~\etal~\cite{lee_CVPRW20_mimicry}.
More details of the training and evaluation settings of GANs appear in the supplementary materials.

\begin{table}[t]
   \centering
       \caption{Mean frequency gaps between real and fake images}
       \label{tb:frequency_gap}
      \resizebox{\columnwidth}{!}{
       \begin{tabular}{lcccccc}\toprule
         & CIFAR-10 & CIFAR-100 & TinyImageNet & STL-10 & CelebA & ImageNet \\
         \midrule
           SNGAN             & 6.89 & 7.01 & 9.83 & 4.19 & 4.49 & 4.83 \\
           Binomial~\cite{frankICML20_leveraging_frequency} & 7.85 & 5.83 & 9.96 & 4.30 & 4.74 & 4.55 \\
           SR~\cite{durall_CVPR20_watch_your_upconv} & 6.12 & 6.80 & 9.77 & 3.98 & 4.48 & 5.70 \\
           SSD-GAN~\cite{chen_AAAI21_SSDGAN} & 6.39 & 6.80 & 9.97 & 4.59 & 4.47 & 4.80 \\
           F-Drop            & 5.94 & 6.36 & 9.29 & 3.87 & 4.60 & 5.39 \\
           F-Match           & 4.84 & 4.87 & 7.36 & 4.04 & 4.46 & 4.52 \\
           F-Drop\&Match     & {\bf 3.93} & {\bf 4.16} & {\bf 6.49} & {\bf 3.86} & {\bf 4.43} & {\bf 4.41} \\
           \bottomrule
       \end{tabular}
      }
\end{table}

\begin{figure}[t]
   \centering
   \begin{minipage}{1.0\columnwidth}
      \centering
      \includegraphics[width=1.0\columnwidth]{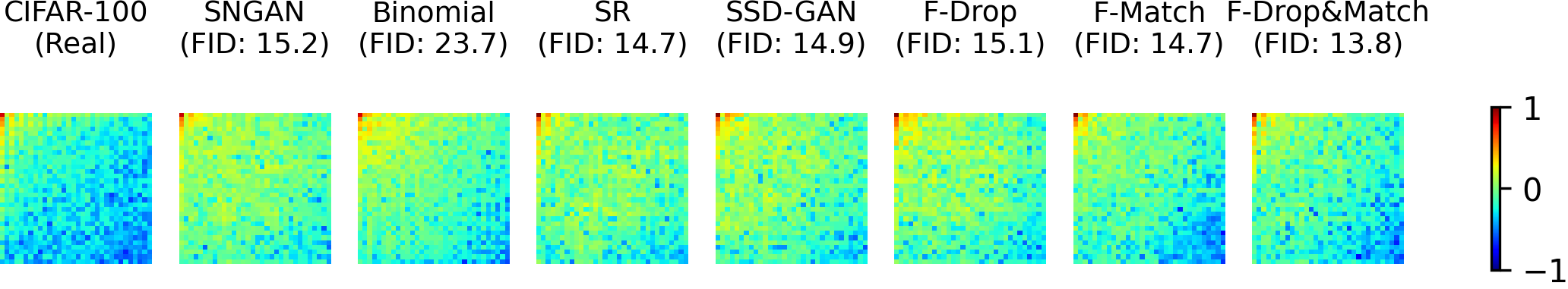}
   \end{minipage}
   \centering
   \begin{minipage}{1.0\columnwidth}
      \centering
      \includegraphics[width=1.0\columnwidth]{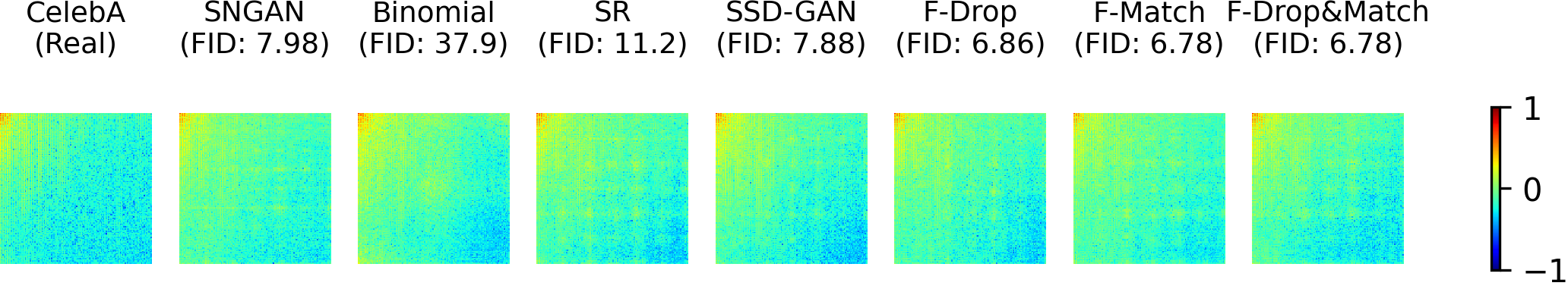}
   \end{minipage}
\caption{
 Comparison of average DCT coefficients (top: CIFAR-100, bottom: CelebA).
 The visualization protocols follow those of Frank~\etal~\cite{frankICML20_leveraging_frequency}.
}
\label{fig:frequency_check}
\end{figure}

\begin{table*}[t]
   \centering
       \caption{Performance comparison on the \(32\times32\) datasets}
       \label{tb:small_data}
       \resizebox{\textwidth}{!}{
       \begin{tabular}{lcccccccccccc}\toprule
         & \multicolumn{3}{c}{CIFAR-10} & \multicolumn{3}{c}{CIFAR-100} & \multicolumn{3}{c}{TinyImageNet} \\
         \cmidrule(lr){2-4}\cmidrule(lr){5-7} \cmidrule(lr){8-10}
         & FID (\(\downarrow\)) & KID{\tiny \(^{\times 10^{-3}}\)} (\(\downarrow\)) & IS (\(\uparrow\)) & FID (\(\downarrow\)) & KID{\tiny \(^{\times 10^{-3}}\)} (\(\downarrow\)) & IS (\(\uparrow\)) & FID (\(\downarrow\)) & KID{\tiny \(^{\times 10^{-3}}\)} (\(\downarrow\)) & IS (\(\uparrow\)) \\
         \midrule
           SNGAN    &  14.3\(^{\pm\text{0.73}}\) & 9.20\(^{\pm\text{0.57}}\) &  8.25\(^{\pm\text{0.14}}\) & 15.2\(^{\pm\text{0.25}}\) & 9.76\(^{\pm\text{0.35}}\) & 8.91\(^{\pm\text{0.04}}\) & 21.8\(^{\pm\text{3.01}}\) & 12.3\(^{\pm\text{3.43}}\) & 6.35\(^{\pm\text{0.28}}\)  \\
           Binomial~\cite{frankICML20_leveraging_frequency}    &  35.9\(^{\pm\text{0.90}}\) & 21.9\(^{\pm\text{1.89}}\) &  6.60\(^{\pm\text{0.15}}\) & 23.7\(^{\pm\text{0.70}}\) & 14.3\(^{\pm\text{0.47}}\) & 8.09\(^{\pm\text{0.08}}\) & 53.9\(^{\pm\text{4.26}}\) & 30.8\(^{\pm\text{8.91}}\) & 5.34\(^{\pm\text{0.29}}\)  \\
           SR~\cite{durall_CVPR20_watch_your_upconv}    &  12.2\(^{\pm\text{0.27}}\) & 7.73\(^{\pm\text{2.73}}\) &  8.43\(^{\pm\text{0.01}}\) & 14.7\(^{\pm\text{0.27}}\) & 9.56\(^{\pm\text{0.49}}\) & 8.94\(^{\pm\text{0.05}}\) & 23.8\(^{\pm\text{2.01}}\) & 18.9\(^{\pm\text{2.00}}\) & 5.96\(^{\pm\text{0.25}}\)  \\
           SSD-GAN~\cite{chen_AAAI21_SSDGAN}    &  13.4$^{\pm\text{0.13}}$ & 8.72$^{\pm\text{0.21}}$ &  8.32$^{\pm\text{0.11}}$ & 14.9$^{\pm\text{0.88}}$ & 9.32$^{\pm\text{0.78}}$ & 9.01$^{\pm\text{0.31}}$ & 21.1$^{\pm\text{1.51}}$ & 12.9$^{\pm\text{1.72}}$ & 6.50$^{\pm\text{0.23}}$  \\
           F-Drop    &  14.1$^{\pm\text{0.81}}$ & 9.11$^{\pm\text{0.21}}$ &  8.31$^{\pm\text{0.18}}$ & 15.1$^{\pm\text{0.15}}$ & 9.47$^{\pm\text{0.29}}$ & 8.93$^{\pm\text{0.05}}$ & 20.4$^{\pm\text{0.46}}$ & 11.5$^{\pm\text{1.18}}$ & 6.49$^{\pm\text{0.08}}$  \\
           F-Match    &  12.8$^{\pm\text{0.53}}$ & 7.90$^{\pm\text{0.32}}$ &  8.45$^{\pm\text{0.12}}$ & 14.7$^{\pm\text{0.66}}$ & 9.09$^{\pm\text{0.89}}$ & 9.17$^{\pm\text{0.24}}$ & 20.9$^{\pm\text{0.24}}$ & 12.7$^{\pm\text{0.46}}$ & 6.41$^{\pm\text{0.24}}$  \\
           F-Drop\&Match    &  {\bf 10.7}$^{\pm\textbf{0.92}}$ & {\bf 7.15}$^{\pm\textbf{0.58}}$ &  {\bf 8.45}$^{\pm\textbf{0.06}}$ & {\bf 13.8}$^{\pm\textbf{0.34}}$ & {\bf 8.99}$^{\pm\textbf{0.49}}$ & {\bf 9.16}$^{\pm\textbf{0.00}}$ & {\bf 18.9}$^{\pm\textbf{1.08}}$ & {\bf 10.3}$^{\pm\textbf{0.52}}$ & {\bf 6.55}$^{\pm\textbf{0.14}}$  \\
           \bottomrule
       \end{tabular}
       }
\end{table*}
\begin{table*}[t]
   \centering
   \caption{Performance comparison on larger image datasets}
   \label{tb:large_data}
   \resizebox{\textwidth}{!}{
      \begin{tabular}{lcccccccccccc}\toprule
         & \multicolumn{3}{c}{STL-10 (\(48\times 48\))} & \multicolumn{3}{c}{CelebA (\(128\times 128\))} & \multicolumn{3}{c}{ImageNet (\(128\times 128\))} \\
         \cmidrule(lr){2-4}\cmidrule(lr){5-7} \cmidrule(lr){8-10}
         & FID (\(\downarrow\)) & KID{\tiny $^{\times 10^{-3}}$} (\(\downarrow\)) & IS (\(\uparrow\)) & FID (\(\downarrow\)) & KID{\tiny $^{\times 10^{-3}}$} (\(\downarrow\)) & IS (\(\uparrow\)) & FID (\(\downarrow\)) & KID{\tiny $^{\times 10^{-3}}$} (\(\downarrow\)) & IS (\(\uparrow\)) \\         
         % \cmidrule(lr){2-4}\cmidrule(lr){5-7} \cmidrule(lr){8-10}
         \midrule
         SNGAN    &  34.7$^{\pm\text{1.26}}$ & 32.0$^{\pm\text{0.91}}$ &  8.68$^{\pm\text{0.08}}$ & 7.98$^{\pm\text{0.13}}$ & 4.45$^{\pm\text{0.42}}$ & 3.02$^{\pm\text{0.07}}$ & 62.5$^{\pm\text{1.16}}$ & 63.5$^{\pm\text{0.80}}$ & 14.1$^{\pm\text{0.34}}$  \\
         Binomial~\cite{frankICML20_leveraging_frequency}    &  34.9$^{\pm\text{0.44}}$ & 32.4$^{\pm\text{1.01}}$ &  8.66$^{\pm\text{0.09}}$ & 37.9$^{\pm\text{6.57}}$ & 22.3$^{\pm\text{0.95}}$ & 2.86$^{\pm\text{0.02}}$ & 76.6$^{\pm\text{6.93}}$ & 74.1$^{\pm\text{6.15}}$ & 11.6$^{\pm\text{1.14}}$  \\
         SR~\cite{durall_CVPR20_watch_your_upconv} & 38.1$^{\pm\text{0.74}}$ & 34.9$^{\pm\text{0.87}}$ &  8.49$^{\pm\text{0.02}}$ & 11.2$^{\pm\text{0.74}}$ & 5.67$^{\pm\text{0.87}}$ & 2.91$^{\pm\text{0.06}}$ & 64.0$^{\pm\text{1.52}}$ & 64.9$^{\pm\text{2.35}}$ & 13.9$^{\pm\text{0.49}}$  \\
         SSD-GAN~\cite{chen_AAAI21_SSDGAN}    &  35.6$^{\pm\text{0.25}}$ & 32.2$^{\pm\text{0.68}}$ &  8.77$^{\pm\text{0.03}}$ & 7.88$^{\pm\text{0.64}}$ & 4.20$^{\pm\text{0.97}}$ & 3.05$^{\pm\text{0.07}}$ & 61.2$^{\pm\text{0.49}}$ & 61.6$^{\pm\text{1.69}}$ & 14.3$^{\pm\text{0.08}}$  \\
         F-Drop    &  34.7$^{\pm\text{0.75}}$ & 31.8$^{\pm\text{1.09}}$ &  8.75$^{\pm\text{0.07}}$ & 6.86$^{\pm\text{0.47}}$ & 3.92$^{\pm\text{0.64}}$ & 3.09$^{\pm\text{0.06}}$ & 61.0$^{\pm\text{0.59}}$ & 60.9$^{\pm\text{1.86}}$ & 14.2$^{\pm\text{0.21}}$  \\
         F-Match    &  34.0$^{\pm\text{0.72}}$ & 31.1$^{\pm\text{0.76}}$ &  8.79$^{\pm\text{0.05}}$ & 6.78$^{\pm\text{0.16}}$ & 3.73$^{\pm\text{0.18}}$ & 3.08$^{\pm\text{0.04}}$ & 62.0$^{\pm\text{1.33}}$ & 62.2$^{\pm\text{1.35}}$ & 14.4$^{\pm\text{0.18}}$  \\
         F-Drop\&Match    &  {\bf 33.8}$^{\pm\textbf{0.66}}$ & {\bf 30.4}$^{\pm\textbf{0.83}}$ &  {\bf 8.85}$^{\pm\textbf{0.15}}$ &{\bf 6.78}$^{\pm\textbf{0.11}}$ & {\bf 3.61}$^{\pm\textbf{0.10}}$ & {\bf 3.16}$^{\pm\text{0.05}}$ & {\bf 60.4}$^{\pm\textbf{0.71}}$ & {\bf 60.5}$^{\pm\textbf{0.51}}$ & {\bf 14.5}$^{\pm\textbf{0.30}}$  \\
         \bottomrule
      \end{tabular}
       }
   \end{table*}
   \begin{table}[!t]
      \centering
          \caption{Performance comparison on GAN variants (CIFAR-100)}
          \label{tb:gan_variants}
          \resizebox{\columnwidth}{!}{
            \begin{tabular}{lcccccc}\toprule
            & \multicolumn{3}{c}{DCGAN~\cite{Radford_ICLR16_DCGAN}} & \multicolumn{3}{c}{WGAN-GP~\cite{gulrajani17_wgangp}} \\
            \cmidrule(lr){2-4}\cmidrule(lr){5-7}
            & FID (\(\downarrow\)) & KID{\tiny $^{\times 10^{-3}}$} (\(\downarrow\)) & IS (\(\uparrow\)) & FID (\(\downarrow\)) & KID{\tiny $^{\times 10^{-3}}$} (\(\downarrow\)) & IS (\(\uparrow\)) \\  
            % \cmidrule(lr){2-4}\cmidrule(lr){5-7} \cmidrule(lr){8-10}
            \midrule
              Baseline    &  27.2$^{\pm\text{1.15}}$ & 16.2$^{\pm\text{1.69}}$ &  7.16$^{\pm\text{0.26}}$ & 25.2$^{\pm\text{0.20}}$ & 21.2$^{\pm\text{0.33}}$ & 7.72$^{\pm\text{0.03}}$ \\
              Binomial~\cite{frankICML20_leveraging_frequency}    &  49.8$^{\pm\text{3.78}}$ & 28.6$^{\pm\text{2.38}}$ & 6.28$^{\pm\text{0.18}}$ & 25.1$^{\pm\text{0.48}}$ & 19.8$^{\pm\text{0.77}}$ & 7.65$^{\pm\text{0.07}}$\\
              SR~\cite{durall_CVPR20_watch_your_upconv}    &  40.8$^{\pm\text{2.28}}$ & 24.4$^{\pm\text{3.07}}$ &  6.24$^{\pm\text{0.27}}$ & 40.9$^{\pm\text{3.68}}$ & 22.7$^{\pm\text{2.84}}$ & 6.28$^{\pm\text{0.45}}$ \\
              SSD-GAN~\cite{chen_AAAI21_SSDGAN}    &  34.2$^{\pm\text{1.58}}$ & 18.6$^{\pm\text{1.62}}$ &  6.53$^{\pm\text{0.23}}$ & 45.0$^{\pm\text{1.76}}$ & 30.1$^{\pm\text{1.96}}$ & 6.04$^{\pm\text{0.32}}$ \\
              F-Drop    &  25.9$^{\pm\text{0.45}}$ & 15.8$^{\pm\text{0.27}}$ &  7.15$^{\pm\text{0.05}}$ & 23.8$^{\pm\text{0.28}}$ & 19.3$^{\pm\text{0.52}}$ & 7.85$^{\pm\text{0.01}}$ \\
              F-Match    &  26.5$^{\pm\text{0.73}}$ & 16.3$^{\pm\text{1.06}}$ &  7.23$^{\pm\text{0.17}}$ & 24.9$^{\pm\text{0.14}}$ & 20.2$^{\pm\text{0.21}}$ & 7.59$^{\pm\text{0.11}}$ \\
              F-Drop\&Match    &  {\bf 25.2}$^{\pm\textbf{1.17}}$ & {\bf 15.4}$^{\pm\textbf{0.44}}$ & {\bf 7.45}$^{\pm\textbf{0.07}}$ & {\bf 23.9}$^{\pm\textbf{0.53}}$ & {\bf 18.9}$^{\pm\textbf{0.79}}$ & {\bf 7.97}$^{\pm\textbf{0.02}}$ \\
              \bottomrule
          \end{tabular}
          }
   \end{table}

\subsection{Main Evaluations}\label{sec:main_result}
\paragraph{Frequency Gaps}
First, we evaluate the reduction in the frequency gaps.
The following total absolute difference in DCT space was used as a measure of the frequency gap:
\begin{equation}\label{eq:frequency_gap}
   \frac{1}{HW}\sum^H_u\sum^W_v\left|\bar{X}_{\text{real}}(u,v) - \bar{X}_{\text{fake}}(u,v) \right|,
\end{equation}
where \(\bar{X}(u,v)\) is defined in Eq.~(\ref{eq:fmatch}).
We computed the gaps between 10k real and generated images, where the real images were randomly selected from each dataset.
Table~\ref{tb:frequency_gap} lists the mean frequency gaps measured by Eq.~(\ref{eq:frequency_gap}).
The F-Drop\&Match column represents the performances of SNGAN simultaneously applying F-Drop and F-Match.
Visualizations of the frequency characteristics are shown in Fig.~\ref{fig:frequency_check}, where the pixels in the upper left represent lower frequency components and ones in the lower right represent higher frequency components.
The figure and table show that F-Drop\&Match significantly reduced the frequency gaps in all datasets and replicated more realistic frequency characteristics compared with the other methods.
In a few cases, F-Drop by itself did not reduce the gaps.
This is because F-Drop allows the generators to synthesize the filtered out high-frequency components, and thus, the generated images contain high-frequency components at random.
On the other hand, F-Match by itself reduced the gaps in all cases, since it directly minimizes the frequency characteristics.
In Fig.~\ref{fig:frequency_check}, the results of F-Match show frequency gaps in the middle range of the frequency domain more so than the results of F-Drop\&Match.
This is because the generator of F-Match (by itself) focuses on high-frequency components because of the sensitivity of the discriminators to the high-frequency domain.
These results indicate that F-Drop\&Match reduces the frequency gaps by complementarily combining filtering and direct minimization.
Furthermore, F-Drop\&Match outperformed the other frequency-oriented methods, \ie, Binomial, SR, and SSD-GAN.
The poorer performance of these other methods is probably because they do not take account of the sensitivity of the discriminators to the high-frequency domain. 
The sensitivity of the other methods is discussed in Sec.~\ref{sec:eval_sfa}.
Here, we discuss other reasons why Binomial, SR, and SSD-GAN are inferior to our methods.
In the case of Binomial, the binomial upsampling suppresses the high-frequency components in the generators, but does not explicitly regularizes the models to learn the frequency characteristics.
Moreover, we found that Binomial tends to degrade the generative performance in the spatial domain (see the evaluations described below).
For SR and SSD-GAN, the performance gains sensitively depend on the dataset.
This behavior reflects the 1D approximation with the azimuthal integral, which implicitly assumes that the real frequency characteristics are distributed in a concentric pattern in DFT space.
Since SSD-GAN does not use the gradients from the frequency classifier for updating the GANs, its performance gains may be unstable.
Meanwhile, F-Match directly minimizes the gaps for each frequency component in an end-to-end fashion and performs stably on the various datasets.

\paragraph{FID/KID/IS}
Second, we measured the Fr\'echet inception distance (FID)~\cite{heusel_ttur_nips17}, kernel inception distance (KID)~\cite{binkowski_ICLR18_KID}, and inception score (IS)~\cite{Salimans_NIPS16}.
We computed these measures on 100k real and generated images for the \(128\!\times\!128\) datasets, and 50k real and generated images for the \(32\!\times\!32\) datasets and STL-10.
Table~\ref{tb:small_data}, \ref{tb:large_data}, and~\ref{tb:gan_variants} show the scores of FID/KID/IS for each combination of dataset, method, and GAN variant.
Note that \(\downarrow\) means lower is better and \(\uparrow\) means higher is better.
F-Drop by itself and F-Match by itself outperformed the baselines in many cases.
More importantly, F-Drop\&Match performed the best in all cases.
Binomial underperformed the baseline in almost all cases.
Similar to the evaluation of the frequency gaps, the performances of SR and SSD-GAN sensitively depend on the datasets, while our methods stably outperform the baselines.
These results indicate that our methods can flexibly help GANs to replicate the real images in both the frequency and spatial domain.

\subsection{Frequency Sensitivity Analysis}\label{sec:eval_sfa}
\begin{figure}
   \centering
         \includegraphics[width=1\columnwidth]{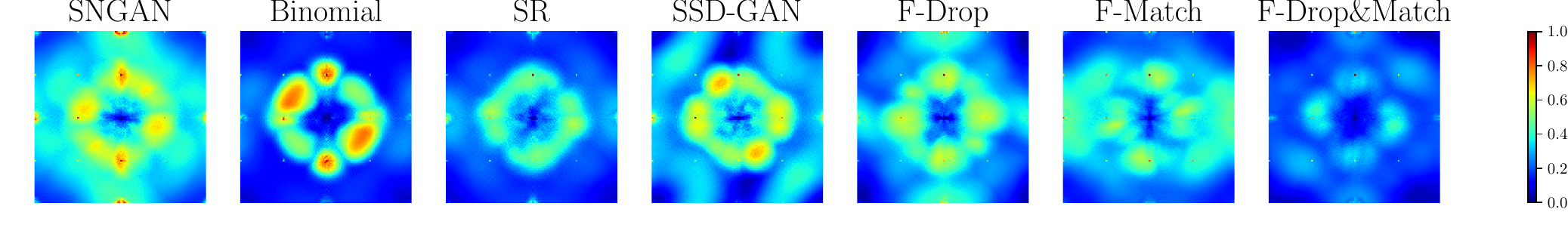}
   \caption{
         Sensitivity analysis by SFA~\cite{tsuzuku_CVPR2019_structural_sensitivity} on CelebA
         }
\label{fig:sfa_celeba}
\end{figure}

\begin{figure}
   \centering
   \begin{tabular}{l}
      \rotatebox[origin=c]{90}{\tiny SNGAN} 
      \begin{minipage}{0.95\columnwidth}
         \centering
         \includegraphics[width=1.0\columnwidth]{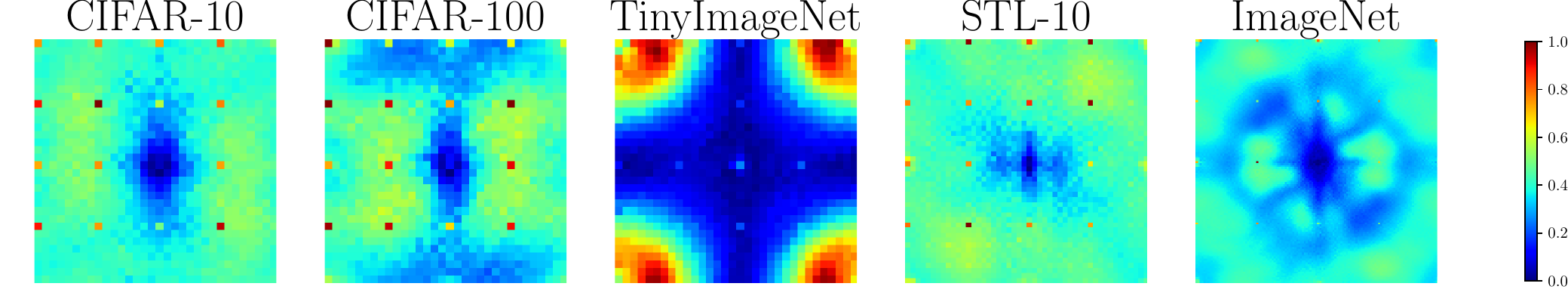} 
      \end{minipage} \\
      \rotatebox[origin=c]{90}{\tiny F-Drop\&Match} 
      \begin{minipage}{0.95\columnwidth}
         \centering
         \includegraphics[width=1.0\columnwidth]{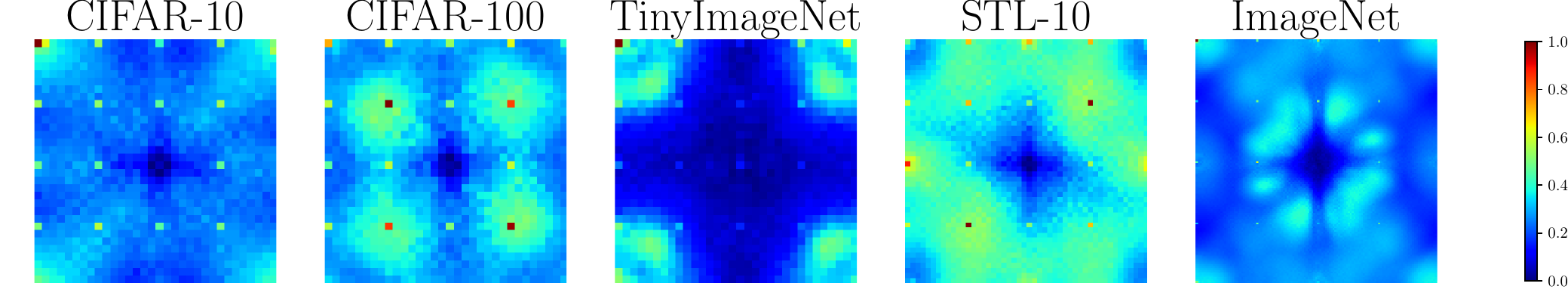}
      \end{minipage} \\
   \end{tabular}
   \caption{
         Sensitivity analysis by SFA~\cite{tsuzuku_CVPR2019_structural_sensitivity} on multiple datasets
         }
   \label{fig:sfa_all}
\end{figure}

As shown in Sec.~\ref{sec:sfa_analysis}, the discriminators of GANs are sensitive to the perturbations in the high-frequency domain.
We evaluate the sensitivity of our methods by conducting an SFA analysis.
Figure~\ref{fig:sfa_celeba} compares the baselines and our methods in terms of the results of SFA perturbing each frequency component on CelebA.
Figure~\ref{fig:sfa_all} shows the results of SFA on multiple datasets.
We used the same visualization protocol as in Sec.~\ref{sec:sfa_analysis}.
We also tested Binomial, SR, and SSD-GAN on CelebA (Fig.~\ref{fig:sfa_celeba}); the results on the other datasets appear in the supplementary materials and lead to the same conclusions as given below.
F-Drop\&Match outperformed all of the baselines.
Since the robust frequency domains of F-Drop\&Match seem to be the union of the robust frequency domains of F-Drop and ones of F-Match, we see that the robustness of F-Drop\&Match comes from combining F-Drop and F-Match complementarily.
More importantly, in Fig.~\ref{fig:sfa_celeba}, we see that Binomial, SR, and SSD-GAN are not robust against the low to middle range of the frequency domain.
We consider that this is because these methods, unlike F-Drop, feed the whole input images including their high-frequency components into the discriminators, and thus, the discriminators have trouble focusing on the lower frequency domain.
These results suggest that the discriminators of F-Drop\&Match can focus on the lower frequency and the generators indirectly adjust their training to learn realistic frequency components by combining F-Drop and F-Match complementarily.


\subsection{Evaluation on StyleGAN2-ADA}\label{sec:stylegan}
\begin{table}[t]
   \centering
       \caption{Evaluation on StyleGAN2-ADA (FID)}
       \label{tb:stylegan}
       \scalebox{0.85}{
       \begin{tabular}{lcccc}\toprule
         & FFHQ & Cat & Dog & Wild \\
         & {\small \((256\times 256)\)} & \multicolumn{3}{c}{\small (\(512\times 512\))} \\
         \midrule
           StyleGAN2-ADA~\cite{karras_arxiv20_training_GANs_with_limited_data} & 4.30 & 3.55 & 7.40 & 3.05 \\
           F-Drop\&Match       & {\bf 4.05} &  {\bf 3.36} & {\bf 7.21} & {\bf 2.62} \\
           \bottomrule
       \end{tabular}
       }
\end{table}
Here, we show extra results evaluated on state-of-the art settings.
We chose StyleGAN2-ADA~\cite{karras_arxiv20_training_GANs_with_limited_data} as the baseline and used the implementation provided by the authors.\footnote{https://github.com/NVlabs/stylegan2-ada-pytorch}
We evaluated models on the four high-resolution datasets: FFHQ~\cite{karras_CVPR19_stylegan}, and AFHQ-Dog/Cat/Wild~\cite{choi_CVPR20_starganv2}.
The training settings of F-Drop\&Match are shared with that of the previous work~\cite{karras_arxiv20_training_GANs_with_limited_data}; we used {\tt paper256} setting for FFHQ and {\tt paper512} setting for the AFHQ datasets, which are preset in the repository.
The hyperparameters of \(\lambda\) were \(1.0\times 10^{-7}\) for FFHQ and \(1.0\times 10^{-8}\) for the AFHQ datasets.
Table~\ref{tb:stylegan} summarizes the results of FID.
Our method succeeded to improve the baseline.
As the same trend in Section~\ref{sec:eval_sfa}, Figure~\ref{fig:sfa_comp} shows that our method can prevent the discriminators to be fooled by the high-frequency components in input.
Thus, F-Drop\&Match can work well even on high-resolution datasets with state-of-the-art GAN variant.

\subsection{Qualitative Results}
\begin{figure}
   \centering
   \begin{tabular}{l}
      \rotatebox[origin=c]{90}{Real}
      \begin{minipage}{0.8\columnwidth}
         \includegraphics[width=1.0\columnwidth]{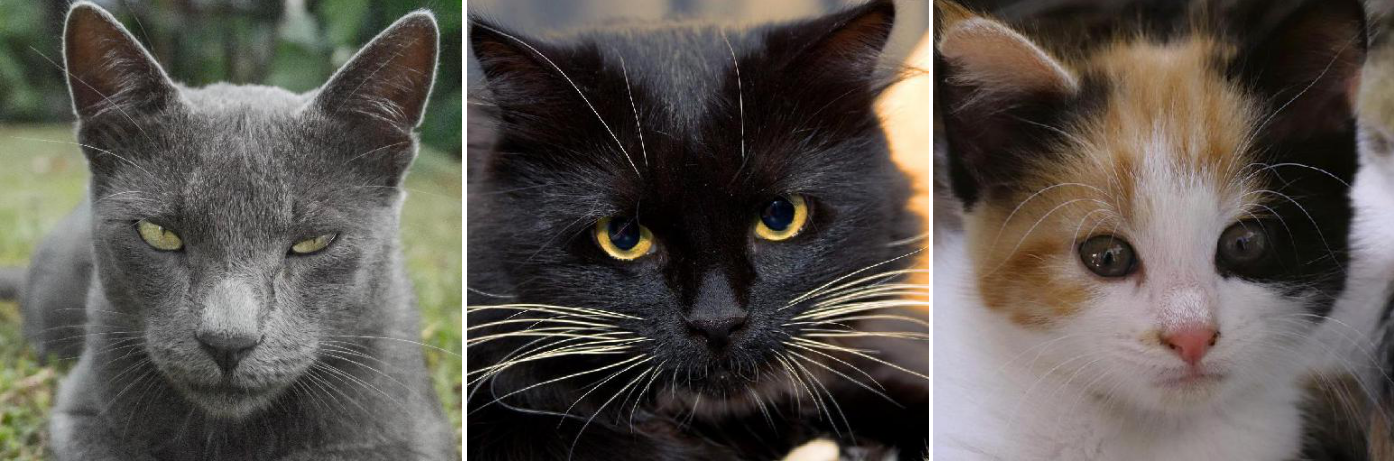} 
      \end{minipage} \\
      \rotatebox[origin=c]{90}{StyleGAN2-ADA}
      \begin{minipage}{0.8\columnwidth}
         \includegraphics[width=1.0\columnwidth]{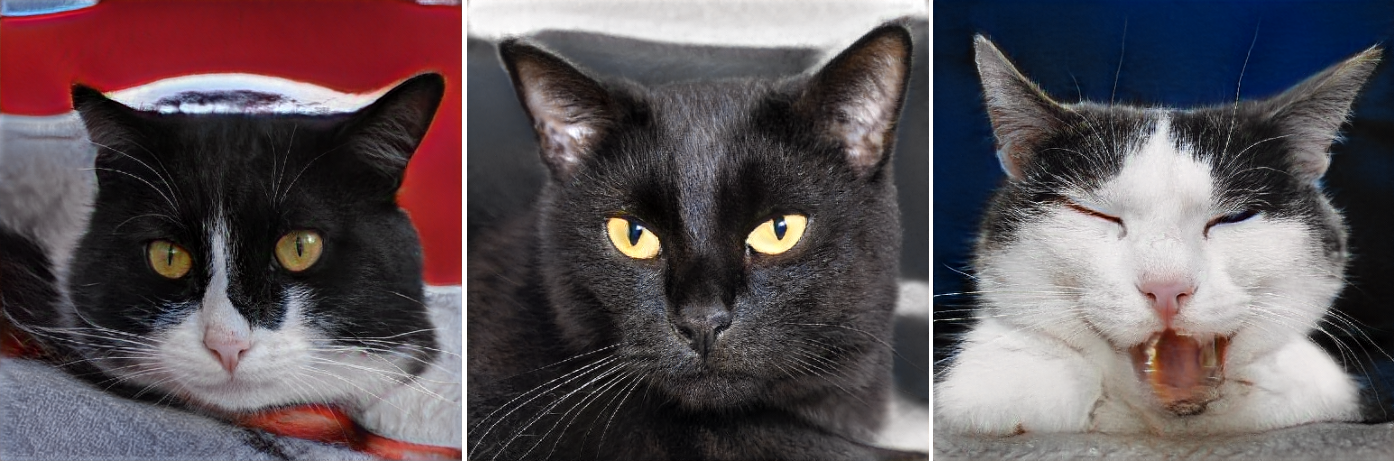}
      \end{minipage} \\
      \rotatebox[origin=c]{90}{Ours}
      \begin{minipage}{0.8\columnwidth}
         \includegraphics[width=1.0\columnwidth]{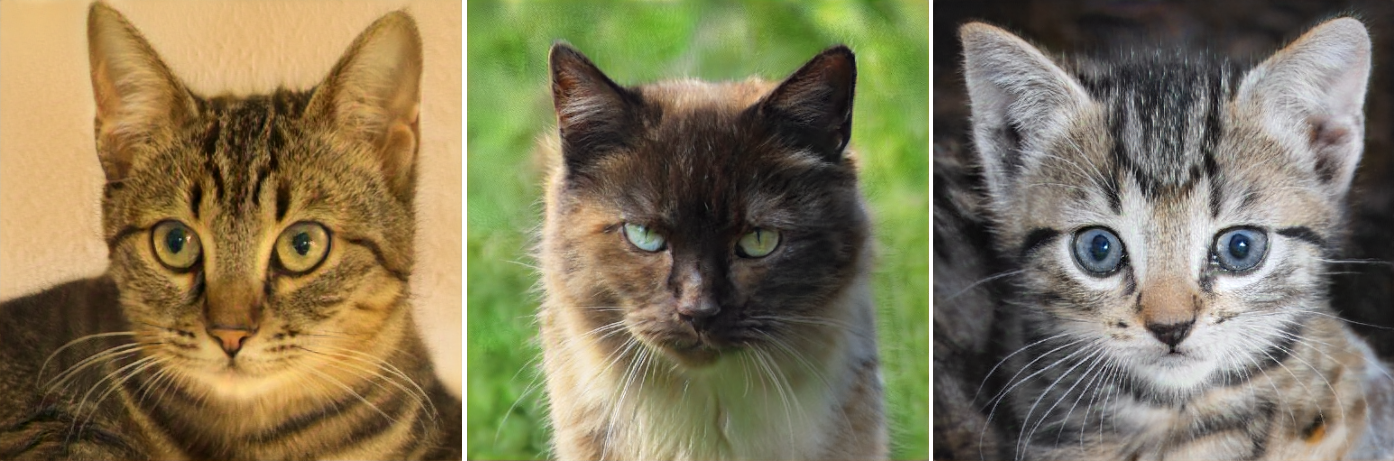}
      \end{minipage} \\
   \end{tabular}
   \caption{
         Visualization of real and generated images (AFHQ-Cat)
         }
   \label{fig:viz_celeba}
\end{figure}

Lastly, we provide visualizations of the generated images.
Figure~\ref{fig:viz_celeba} illustrates the generated images from the StyleGAN2 and F-Drop\&Match (ours) models trained on the AFHQ-Cat dataset; the training settings are shared with Section~\ref{sec:stylegan}.
The generated images are randomly selected.
We emphasize again that we did not use \(\mathbf{M}(\gamma)\) of F-Drop in the evaluation after training.
We can see that both StyleGAN2-ADA and ours synthesized rough shapes of cat faces that are composed of lower frequency components.
Meanwhile, ours was superior to StyleGAN2-ADA at synthesizing more detailed information such as hair which is composed of higher frequency components, while keeping the information on the lower frequency components such as the positions of facial parts.
These results indicate that F-Drop and F-Match make the generators focus on fitting all of frequency components.
More importantly, we found that F-Drop produces no visible flaws by filtering the high-frequency components during training.
Additional visualization studies including ones on other datasets can be found in the supplementary materials; they show the same tendency as described here.

\section{Conclusion}
We presented F-Drop and F-Match for minimizing the frequency gaps that appear in images generated by GANs.
We demonstrated that the discriminators of GANs are highly sensitive to high-frequency perturbations and the sensitivity can cause frequency gaps.
Our methods improve GANs in both the frequency and spatial domain because F-Drop protects the discriminators from high-frequency perturbations and F-Match directly minimizes the frequency gap by using a simple mini-batch error function.
Our extensive experiments show that the combination of F-Drop and F-Match outperforms the baselines on various datasets.
An important direction of future research will be to introduce adaptive masking without any hyperparameter into F-Drop for filtering effectively and generating realistic images.
\clearpage

\appendix
\section*{Supplementary Materials}
This manuscript is the supplementary materials of the main paper (F-Drop\&Match: Improved Techniques for GANs in Frequency Domain).
We provide 
(\ref{sec:dft_vs_dct}) the comparative studies of using DCT and DFT for our methods,
(\ref{sec:training_setting}) the detailed information of training settings,
(\ref{sec:ablation_fmatch}) the ablation studies of F-Match when changing \(d\) and \(\mathcal{F}\) in Eq.~(8) of the main paper,
(\ref{sec:dai}) the implementation details of the differentiable azimuthal integral for spectral regularization (SR)~\cite{durall_CVPR20_watch_your_upconv},
(\ref{sec:hyper}) the detailed settings and sensitively analysis of the hyperparameter \(\gamma\) and \(\lambda\),
(\ref{sec:drop_viz}) the visual effects caused by F-Drop during training,
(\ref{sec:freq_gap_analysis}) the additional analysis of the frequency gaps for confirming the validity of the evaluation and the performances of F-Drop in the lower-frequency domain,
(\ref{sec:fake_detection}) the results of fake image detection evaluation,
(\ref{sec:add_sfa}) the additional sensitivity analysis by single Fourier attack~\cite{tsuzuku_CVPR2019_structural_sensitivity},
(\ref{sec:add_qual}) the additional visualization studies of the images generated from GANs.

\section{Discussion of Frequency Transformations}\label{sec:dft_vs_dct}
In this section, we discuss the reason why we use discrete cosine transform (DCT) for F-Drop and F-Match instead of discrete Fourier transform (DFT) that are used in SR~\cite{durall_CVPR20_watch_your_upconv} and SSD-GAN~\cite{chen_AAAI21_SSDGAN}.

Two-dimensional discrete Fourier transform (DFT) for a squared image \(X \in \mathbb{R}^{H\times H}\) in the spatial domain is defined as:
\begin{equation}
   F(u, v)\!=\!\sum_{i=0}^{H-1} \sum_{j=0}^{H-1} X(i, j) \exp\!\left[ -2 \pi {\rm j} \left( \frac{u i}{H}\!+\!\frac{v j}{H} \right)\!\right],
\end{equation}
where \((i,j)\) represents a spatial pixel coordinate, \((u,v)\) is a frequency coordinate, and \({\rm j}\) is an imaginary unit.
By Euler's formula (\(\exp(\mathrm{j},\theta)=\cos\theta + \mathrm{j}\sin\theta\)), DFT represents an input image with complex values composed of periodic (\ie, sine and cosine functions).
We can translate that DFT treats an input signal as two-dimensional periodic functions represented by extensively tiling the image in the spatial domain.
Thus, DFT produces high-frequency distortions because of the discontinuous boundaries derived from the tiling; this is known as {\em end effects} of DFT~\cite{tribolet_IEEE1979_frequency_end_effect}.
For avoiding the end effects, we use DCT, which does not have the discontinuous boundaries~\cite{tribolet_IEEE1979_frequency_end_effect}.
In contrast to DFT, DCT represents an input image by only cosine functions of real values.
Thus, we can say that DCT treats an input signal as two-dimensional periodic functions represented by symmetrically tiling the image, \ie, DCT does not have the discontinuous boundaries by definition.
We experimentally confirm the performance gaps between DFT and DCT in Sec.~\ref{sec:ablation_fmatch}.

\section{Detailed Training Settings}\label{sec:training_setting}
We basically followed the settings of~\cite{Lee_WACV21_InfoMaxGAN}.
We trained the GANs for 100k iterations on the datasets except for ImageNet (450k iterations on ImageNet).
In all cases, we optimized the GANs with a batch of 64 by using Adam (\(\beta_{1}=0, \beta_{2}=0.9\))~\cite{kingma_adam_iclr14}.
The learning rate of the generators and discriminators was \(2.0 \times 10^{-4}\).
As default settings, we selected \(\gamma=0.8\) for F-Drop by searching in \([0.5, 0.9]\).
For F-Match, we used \(\lambda=1.0\times 10^{-2}\) on the \(32\!\times\!32\) datasets, \(\lambda=1.0\times 10^{-4}\) on STL-10 (\(48\!\times\!48\)), \(\lambda=1.0\times 10^{-5}\) on the \(128\!\times\!128\) datasets; we found them by searching in \([1.0\times 10^{-6}, 1.0\times 10^1]\).
The supplementary materials provide details on the hyperparameter search settings.
In all experiments, we trained GANs three times, and show the mean and standard deviation of each metric.
We evaluated the Fr\'chet inception distance (FID) after 1k iterations and picked the best FID model.
Note that we did not use \(\mathbf{M}(\gamma)\) of F-Drop in the evaluations conducted after training.

\section{Ablation Study of F-Match}\label{sec:ablation_fmatch}
Here, we provide the ablation study for F-Match testing the multiple combinations of the error function \(d(\cdot)\) and the frequency transformation \(\mathcal{F}(\cdot)\) (\eg, DFT and DCT) in Eq.~(8) of the main paper.
We basically share the settings of training and network architectures with Section 6 of the main paper.

\begin{table}[t]
   \centering
         \caption{Comparison among F-Match family (CIFAR-100)}
         \label{tb:ablation_match}
      \resizebox{\columnwidth}{!}{
      \begin{tabular}{lccc}\toprule
         & FID (\(\downarrow\)) & KID{\tiny $^{\times 10^{-3}}$} (\(\downarrow\)) & IS (\(\uparrow\))\\
         % \cmidrule(lr){2-4}\cmidrule(lr){5-7} \cmidrule(lr){8-10}
         \midrule
            Baseline (SNGAN)   &  15.2$^{\pm\text{0.25}}$ & 9.76$^{\pm\text{0.35}}$ & 8.91$^{\pm\text{0.04}}$ \\
            MSE (Pixel)  &  15.3$^{\pm\text{0.26}}$ & 9.67$^{\pm\text{0.29}}$ & 8.99$^{\pm\text{0.12}}$\\
            MSE (DFT)  &  15.0$^{\pm\text{0.36}}$ & 9.20$^{\pm\text{0.24}}$ & 9.06$^{\pm\text{0.10}}$\\
            MSE (DCT)  &  {\bf 14.7}$^{\pm\textbf{0.66}}$ & {\bf 9.09}$^{\pm\textbf{0.89}}$ & {\bf 9.17}$^{\pm\textbf{0.24}}$\\
            MAE (DCT)  &  14.9$^{\pm\text{0.07}}$ & 9.40$^{\pm\text{0.84}}$ & 9.01$^{\pm\text{0.00}}$\\
            MKL (DCT)  &  15.5$^{\pm\text{0.24}}$ & 9.89$^{\pm\text{0.05}}$ & 9.01$^{\pm\text{0.10}}$\\
            MSSE (DCT) &  14.8$^{\pm\text{0.23}}$ & 9.17$^{\pm\text{0.41}}$ & 9.12$^{\pm\text{0.11}}$\\
            \bottomrule
         \end{tabular}
         }
\end{table}

As defined in Eq.~(8) of the main paper, F-Match can equip arbitrary error function \(d\) and frequency transforms \(\mathcal{F}\).
We explore multiple combinations of \(d\) and \(\mathcal{F}\) for F-Match.
We tested DFT, DCT and Pixel (identity function) as \(\mathcal{F}\) and the following four error functions as \(d\): MSE, mean absolute error (MAE), mean KL-divergence (MKL), MSE with concatenating mean and standard deviation of batch frequency components (MSSE).
In Table~\ref{tb:ablation_match}, we summarize the ablation study of F-Match.
Among the variations, MSE in DCT spaces achieved the best performance in terms of FID/KID/IS.
We confirm that minimizing the gap in the frequency domain by using DFT or DCT helps boost the generative performance of GANs whereas minimizing the gap in the spatial domain (Pixel) does not change the performance.
In comparison among frequency transforms, DCT is superior to DFT as we expected in Sec.~\ref{sec:dft_vs_dct}.
Further, in comparison among error functions, we confirm MSE is the best choice.

\section{Differentiable Azimuthal Integral}\label{sec:dai}
In the main paper, we used the differentiable version of spectral regularization (SR).
We reimplemented the differentiable SR with PyTorch because the original reproduction code of azimuthal integral that is published by the author of~\cite{durall_CVPR20_watch_your_upconv} was implemented by Numpy, \ie, it was not differentiable.\footnote{https://github.com/cc-hpc-itwm/UpConv}
For confirming the validity of the reimplementation, we show the reimplementation code of the differentiable azimuthal integral and the comparison results of the non-differentiable and differentiable versions.
The reimplementation code was basically constructed by replacing the Numpy functions in the original code with the corresponding PyTorch functions.
We tested the performances with SNGAN and CIFAR-100 as well as Section 6 of the main paper.
We used the original code of~\cite{durall_CVPR20_watch_your_upconv} as the non-differentiable version.
Algorithm~\ref{alg:azimuthal_integral} shows the code and Table~\ref{tb:sr} lists the performance comparison.
In Table~\ref{tb:sr}, our differentiable SR succeeded to outperform the baseline whereas the non-differentiable SR did not.
This result suggests that our reimplementation has a certain validity.

\begin{algorithm}[t]
   \caption{Azimuthal Integral in PyTorch}
   \label{alg:azimuthal_integral}
   \definecolor{codeblue}{rgb}{0.25,0.5,0.5}
   \lstset{
     backgroundcolor=\color{white},
     basicstyle=\fontsize{7.2pt}{7.2pt}\ttfamily\selectfont,
     columns=fullflexible,
     breaklines=true,
     captionpos=b,
     commentstyle=\fontsize{7.2pt}{7.2pt}\color{codeblue},
     keywordstyle=\fontsize{7.2pt}{7.2pt},
   %  frame=tb,
   }
   \begin{lstlisting}[language=python]
def azimuthal_integral(fft_image, center=None):
   # Calculate the indices from the image
   # These indices are ok to be numpy array
   x, y = np.indices(list(fft_image.shape))
   x, y = torch.from_numpy(x).cuda(), torch.from_numpy(y).cuda()

   if not center:
         center = torch.tensor([(x.max() - x.min()) / 2.0, (y.max() - y.min()) / 2.0])

   r = torch.hypot(x - center[0], y - center[1])

   # Get sorted radii
   ind = torch.argsort(r.flatten())
   r_sorted = r.flatten()[ind]
   i_sorted = fft_image.flatten()[ind]

   # Get the integer part of the radii (bin size = 1)
   r_int = r_sorted.int()

   # Find all pixels that fall within each radial bin.
   deltar = r_int[1:] - r_int[:-1]
   rind = torch.where(deltar)[0]
   nr = rind[1:] - rind[:-1]

   # Cumulative sum to figure out sums for each radius bin
   csim = torch.cumsum(i_sorted, dim=0, dtype=torch.float32)
   tbin = csim[rind[1:]] - csim[rind[:-1]]

   radial_prof = tbin / nr

   return radial_prof
   \end{lstlisting}
   \end{algorithm}

\begin{table}[t]
   \centering
         \caption{Comparison of differentiable and non-differentiable implementation of SR (CIFAR-100)  }
         \label{tb:sr}
      \resizebox{\columnwidth}{!}{
      \begin{tabular}{lccc}\toprule
         & FID (\(\downarrow\)) & KID{\tiny $^{\times 10^{-3}}$} (\(\downarrow\)) & IS (\(\uparrow\))\\
         \midrule
            Baseline (SNGAN)   &  15.2$^{\pm\text{0.25}}$ & 9.76$^{\pm\text{0.35}}$ & 8.91$^{\pm\text{0.04}}$ \\
            Non-Differentiable SR  &  15.8$^{\pm\text{0.11}}$ & 9.81$^{\pm\text{0.54}}$ & 8.85$^{\pm\text{0.09}}$ \\
            Differentiable SR (our reimpl.)  &  14.7$^{\pm\text{0.27}}$ & 9.56$^{\pm\text{0.49}}$ & 8.94$^{\pm\text{0.05}}$ \\
            \bottomrule
         \end{tabular}
         }
\end{table}

\section{Details of Hyperparameter Search}\label{sec:hyper}
\begin{figure}[t]
   \centering
   \includegraphics[width=1.0\columnwidth]{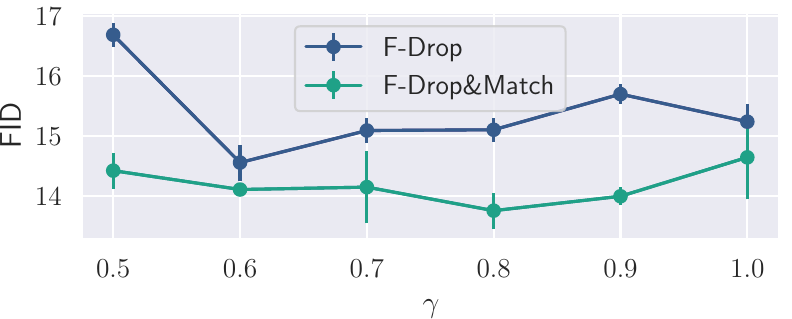}
   \caption{
      Effect of hyperparameter \(\gamma\) in F-Drop (CIFAR-100)
   }
   \label{fig:r_drop}
\end{figure}
\begin{figure}[t]
   \centering
   \includegraphics[width=1.0\columnwidth]{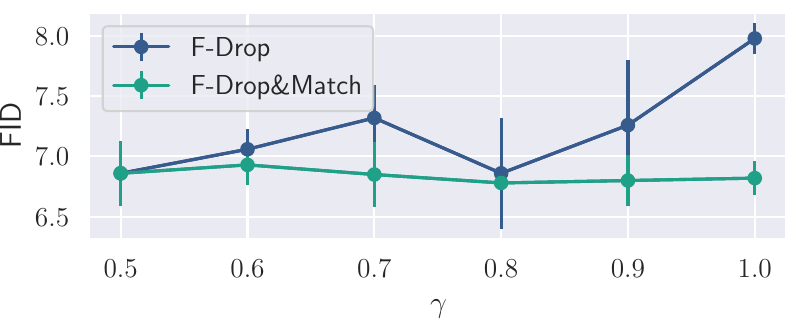}
   \caption{
      Effect of hyperparameter \(\gamma\) in F-Drop (CelebA)
   }
   \label{fig:r_drop_celeba}
\end{figure}
\begin{figure}[t]
   \centering
   \includegraphics[width=1.0\columnwidth]{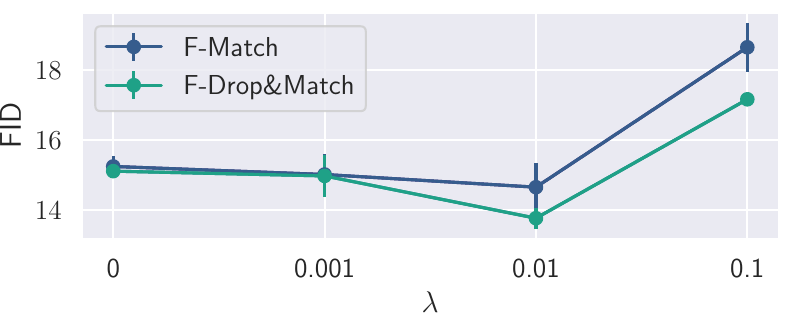}
   \caption{
      Effect of hyperparameter \(\lambda\) in F-Match (CIFAR-100)
   }
   \label{fig:lambda_match}
\end{figure}
\begin{figure}[t]
   \centering
   \includegraphics[width=1.0\columnwidth]{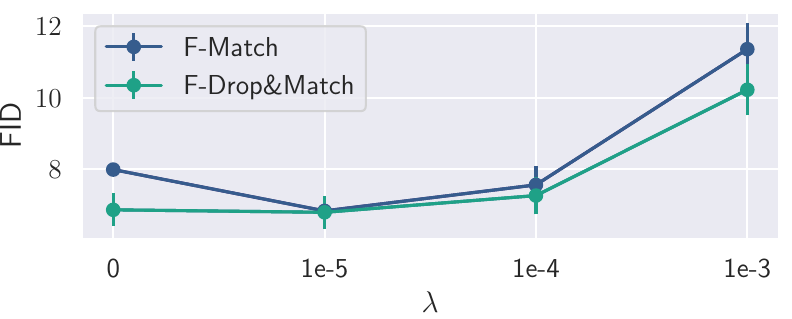}
   \caption{
      Effect of hyperparameter \(\lambda\) in F-Match (CelebA)
   }
   \label{fig:lambda_match_celeba}
\end{figure}
In this section, we describe the details of the hyperparameter search of \(\gamma\) and \(\lambda\) in F-Drop and F-Match.
We also show the sensitivity analysis when changing the hyperparameters.

For \(\gamma\), we searched the values in \(\{0.5, 0.6, 0.7, 0.8, 0.9\}\) with SNGAN on CIFAR-100.
Fig.~\ref{fig:r_drop} illustrates the sensitivity to \(\gamma\) (\(\gamma = 1.0\) means the baseline models).
In both CIFAR-100 and CelebA, the best \(\gamma\) was \(0.8\) for F-Drop\&Match.
The models of F-Drop were inferior to the baselines (SNGANs) in some cases.
This is because the generators of F-Drop synthesize filtered out high-frequency components at random, and thus, the high-frequency components may prevent the training.
In contrast, the F-Drop\&Match models stably outperformed the baselines and F-Drop models with the same \(\gamma\).
Furthermore, we can confirm that there is a difference between single F-Drop and F-Drop\&Match in the tendencies; the best \(\gamma\) were 0.5 or 0.6 for F-Drop by itself and 0.8 for F-Drop\&Match.
This implies that F-Match helps generators to synthesize more realistic high-frequency components that were not learned well by the F-Drop models with \(\gamma=0.8\).

For \(\lambda\), we searched the values in \(\{1.0\times 10^{-6}, 1.0\times 10^{-5}, 1.0\times 10^{-5}, 1.0\times 10^{-4}, 1.0\times 10^{-3}, 1.0\times 10^{-2}, 1.0\times 10^{-1}, 1.0\times 10^{0}, 1.0\times 10^{1}\}\) with SNGAN on each dataset.
Fig.~\ref{fig:lambda_match} and~\ref{fig:lambda_match_celeba} illustrate the sensitivity analysis of \(\lambda\) on CIFAR-100 and CelebA (\(\lambda = 0\) means the baseline models).
We can see that the relatively small values contributed to improving the baseline in both single F-Match and F-Drop\&Match.
In contrast to the case of \(\gamma\), the best values of \(\lambda\) are different between CIFAR-100 and CelebA.
The best values of \(\lambda\) highly depend on the resolution of the input images because the scale of the adversarial losses are changed by the logit size of the discriminators that is different by the resolution.
Thus, the best values of \(\lambda\) are transferable across the same resolution datasets (\eg. \(\lambda=1.0\times 10^{-2}\) for \(32\times 32\) datasets and \(\lambda=1.0\times 10^{-5}\) for \(128\times 128\)).

\section{Visual Effects by F-Drop during Training}\label{sec:drop_viz}
\begin{figure*}
   \centering
   \begin{tabular}{lcccccc}
      & \(\gamma=0.0\) & \(\gamma=0.2\) & \(\gamma=0.4\) & \(\gamma=0.6\) & \(\gamma=0.8\) & \(\gamma=1.0\)\\
      \rotatebox[origin=c]{90}{Spatial} &
      \begin{minipage}{0.13\textwidth}
         \centering
         \includegraphics[width=1.0\columnwidth]{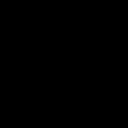} 
      \end{minipage} &
      \begin{minipage}{0.13\textwidth}
         \centering
         \includegraphics[width=1.0\columnwidth]{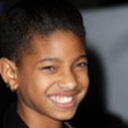} 
      \end{minipage} &
      \begin{minipage}{0.13\textwidth}
         \centering
         \includegraphics[width=1.0\columnwidth]{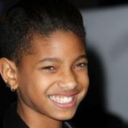} 
      \end{minipage} &
      \begin{minipage}{0.13\textwidth}
         \centering
         \includegraphics[width=1.0\columnwidth]{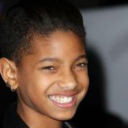} 
      \end{minipage} &
      \begin{minipage}{0.13\textwidth}
         \centering
         \includegraphics[width=1.0\columnwidth]{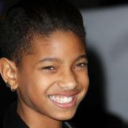} 
      \end{minipage} &
      \begin{minipage}{0.13\textwidth}
         \centering
         \includegraphics[width=1.0\columnwidth]{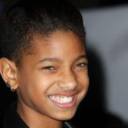} 
      \end{minipage} \\ \\
      \rotatebox[origin=c]{90}{Frequency} &
      \begin{minipage}{0.13\textwidth}
         \centering
         \includegraphics[width=1.0\columnwidth]{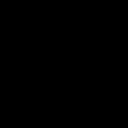} 
      \end{minipage} &
      \begin{minipage}{0.13\textwidth}
         \centering
         \includegraphics[width=1.0\columnwidth]{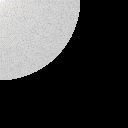} 
      \end{minipage} &
      \begin{minipage}{0.13\textwidth}
         \centering
         \includegraphics[width=1.0\columnwidth]{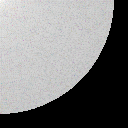} 
      \end{minipage} &
      \begin{minipage}{0.13\textwidth}
         \centering
         \includegraphics[width=1.0\columnwidth]{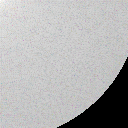} 
      \end{minipage} &
      \begin{minipage}{0.13\textwidth}
         \centering
         \includegraphics[width=1.0\columnwidth]{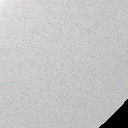} 
      \end{minipage} &
      \begin{minipage}{0.13\textwidth}
         \centering
         \includegraphics[width=1.0\columnwidth]{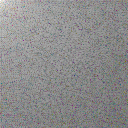} 
      \end{minipage} 
   \end{tabular}
   \caption{Effects of F-Drop on an CelebA sample}
   \label{fig:fdrop}
\end{figure*}

Here, we discuss the visual effects in the spatial domain of input images by applying F-Drop.
Figure~\ref{fig:fdrop} illustrates the effects of F-Drop on the spatial domain and frequency domain when changing the threshold parameter \(\gamma\).
In all cases except for \(\gamma = 0.0\), F-Drop kept most of the spatial information even it filtered out the higher frequency domain.
This indicates that F-Drop does not cause the negative effects during the training of GANs.

\section{Detailed Analysis of Frequency Gaps}\label{sec:freq_gap_analysis}

\begin{table}[t]
   \centering
       \caption{Frequency gaps among real datasets}
       \label{tb:frequency_gap_real}
      \resizebox{\columnwidth}{!}{
       \begin{tabular}{lccccc}\toprule
         & CIFAR-10 & CIFAR-100 & TinyImageNet & CelebA & ImageNet \\
         \midrule
           CIFAR-10      & 2.99 & 3.46 & 4.40 & N/A & N/A\\
           CIFAR-100     & 3.46 & 3.03 & 4.47 & N/A & N/A\\
           TinyImageNet  & 4.40 & 4.47 & 3.24 & N/A & N/A\\
           CelebA        & N/A  & N/A  & N/A  & 2.95 & 3.92 \\
           ImageNet      & N/A  & N/A  & N/A  & 3.92 & 3.21\\
           \bottomrule
       \end{tabular}
      }
\end{table}

\begin{table}[t]
   \centering
       \caption{Frequency gaps in the lower frequency domain}
       \label{tb:frequency_gap_lower}
      \resizebox{\columnwidth}{!}{
       \begin{tabular}{lcccc}\toprule
         & \multicolumn{2}{c}{CIFAR-100} & \multicolumn{2}{c}{CelebA} \\
         \cmidrule(lr){2-3}\cmidrule(lr){4-5}
         & All-band & Lower-band \((\gamma=0.8)\) & All-band & Lower-band \((\gamma=0.8)\) \\
         \midrule
           SNGAN                                            & 7.01 & 5.06 (-1.95) & 4.49 & 4.06 (-0.43)\\
           Binomial~\cite{frankICML20_leveraging_frequency} & 5.83 & 4.55 (-1.28) & 4.74 & 4.22 (-0.52)\\
           SR~\cite{durall_CVPR20_watch_your_upconv}        & 6.80 & 4.75 (-2.05) & 4.48 & 4.22 (-0.26)  \\
           SSD-GAN~\cite{chen_AAAI21_SSDGAN}                & 6.80 & 4.95 (-1.85) & 4.47 & 4.11 (-0.36) \\
           F-Drop                                           & 6.36 & 4.74 (-1.62) & 4.60 & 4.05 (-0.55) \\
           F-Match                                          & 4.87 & 3.97 (-0.90) & 4.46 & 4.04 (-0.42)\\
           F-Drop\&Match                                    & {\bf 4.16} & {\bf 3.80} (-0.36) & {\bf 4.43} & {\bf 3.98} (-0.45) \\
           \bottomrule
       \end{tabular}
      }
\end{table}

We provide additional results of the frequency gaps in terms of (i) the validity of the evaluations by the frequency gaps, and (ii) the comparison of the frequency gaps in the lower frequency domain.

First, we confirm the validity of the measurement of the frequency gaps computed by the mean absolute error defined in Eq.~(14) of the main paper.
To this end, we computed the frequency gaps among the real datasets with respect to the same resolution, \eg, the frequency gaps between CIFAR-10 and TinyImageNet.
Table~\ref{tb:frequency_gap_real} lists the gaps among the real datasets.
We used randomly sampled 10,000 images for each dataset by the same protocol in Sec.~6.2 of the main paper.
Note that we measured the gaps between the same datasets (\eg, CIFAR-10 and CIFAR-10) by using the two different randomly sampled subsets.
The gaps between the real images were in a similar range to the gaps between the real and fake images in Table~1 of the main paper.
Furthermore, we see that F-Drop\&Match can reduce the gaps at the level of the gaps between real images, \eg, in CIFAR-100, 4.16 of F-Drop\&Match is smaller than 4.40 of TinyImageNet.
These results indicate that the mean absolute error is reasonable for measuring the frequency gaps and our method can reduce the gaps to be comparable with the gaps between real datasets.

Next, we show the detailed analysis of the frequency gaps in the lower frequency domain.
In Table~1 of the main paper, we confirm that the models of F-Drop do not reduce the gaps in some cases (\eg, CelebA).
We hypothesize that this is because F-Drop allows the generators to synthesize the filtered out high-frequency components at random.
If this hypothesis is true, the gaps should be reduced when they are measured in the lower-frequency domain without the filtered out high-frequency components.
Table~\ref{tb:frequency_gap_lower} lists the gaps in the lower-frequency domain.
We measured the gap by filtering out the high-frequency components of input images with the mask matrix \(\mathbf{M}(\gamma)\) in Eq.~(7) of the main paper (denoted as Lower-band (\(\gamma=0.8\))).
We used \(\gamma=0.8\) that is the same parameter used in the training of F-Drop by itself and F-Drop\&Match.
The columns of All-band represent the gaps in all frequency band, and they are reprinted from Table~1 of the main paper.
The inside values in the parenthesis of the columns of Low-band are the differences between the Lower-band and All-band values.
The gaps of Lower-band were entirely smaller than that of All-band.
In particular, the Lower-band gaps of F-Drop by itself were significantly reduced from All-band.
Furthermore, we see that F-Drop by itself succeeded in outperforming the baselines in the Lower-band setting.
These results suggest that F-Drop makes GANs concentrate on the training of the lower-frequency components.

\section{Fake Detection}\label{sec:fake_detection}
Similar to the evaluation presented in Frank~\etal~\cite{frankICML20_leveraging_frequency}, we evaluate the detectability of the generated images by using simple linear binary classification models that predict whether an image is real or fake.
By measuring the accuracy of these models, we can assess the quality of the generated images in the spatial and frequency domains.
The input consisted of pixel values or DCT coefficients of the generated images, and the output was a real value in \([0,1]\) representing real or fake.
Similar to~\cite{frankICML20_leveraging_frequency}, we trained the linear regression model with a batch size of 64 by using Adam (\(\beta_{1}=0, \beta_{2}=0.9\), learning rate was 0.001) for 100 epochs.
The real images were taken from the CIFAR-100 dataset and the fake images were generated by each method trained on CIFAR-100.

Table~\ref{tb:deep_fake} lists the mean accuracy of the fake detection models for each setting in CIFAR-100, where Spatial and Frequency represent the results when the pixel values or the DCT coefficients of the generated images are used as the input.
Our methods succeeded in degrading the fake detection accuracy in both the spatial and frequency domain; this means they created more realistic images.
In addition, the Binomial models slightly degraded accuracy compared with the baseline in the frequency domain but improved accuracy in the spatial domain.
This result is consistent with the evaluation in Sec.~6.2 of the main paper: applying a low-pass filter to GAN architectures may lead to difficulty in the training.

\begin{table}[t]
   \centering
       \caption{Mean accuracy of fake detection with linear binary classification (CIFAR-100) }
       \label{tb:deep_fake}
       \scalebox{0.85}{
       \begin{tabular}{lcc}\toprule
         & Spatial & Frequency \\
         \midrule
           Baseline (SNGAN)          &  90.4$^{\pm\text{1.2}}$ & 92.1$^{\pm\text{1.0}}$ \\
           Binomial~\cite{frankICML20_leveraging_frequency} &  95.7$^{\pm\text{0.3}}$ & 90.9$^{\pm\text{0.5}}$ \\
           SR~\cite{durall_CVPR20_watch_your_upconv} &  88.2$^{\pm\text{1.2}}$ & 91.7$^{\pm\text{0.9}}$ \\
           SSD-GAN~\cite{chen_AAAI21_SSDGAN} &  89.7$^{\pm\text{1.6}}$ & 93.2$^{\pm\text{0.6}}$ \\
           F-Drop              &  87.1$^{\pm\text{3.2}}$ & 89.8$^{\pm\text{2.4}}$ \\
           F-Match             &  81.0$^{\pm\text{1.2}}$ & 84.7$^{\pm\text{1.3}}$ \\
           F-Drop\&Match       &  {\bf 78.1}$^{\pm\textbf{0.7}}$ & {\bf 83.1}$^{\pm\textbf{1.9}}$ \\
           \bottomrule
       \end{tabular}
       }
\end{table}

\section{Additional Results of Single Fourier Attack}\label{sec:add_sfa}
\begin{figure*}
   \centering
   \begin{tabular}{l}
      \rotatebox[origin=c]{90}{CIFAR-10} 
      \begin{minipage}{0.95\textwidth}
         \centering
         \includegraphics[width=1.0\columnwidth]{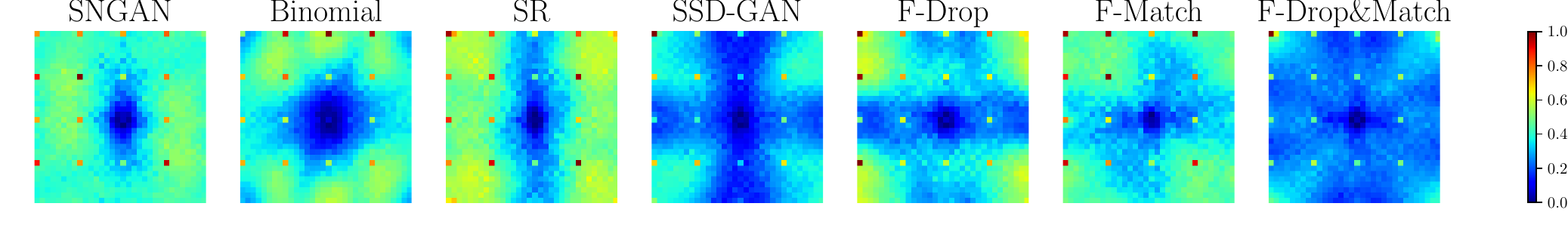} 
      \end{minipage} \\
      \rotatebox[origin=c]{90}{CIFAR-100} 
      \begin{minipage}{0.95\textwidth}
         \centering
         \includegraphics[width=1.0\columnwidth]{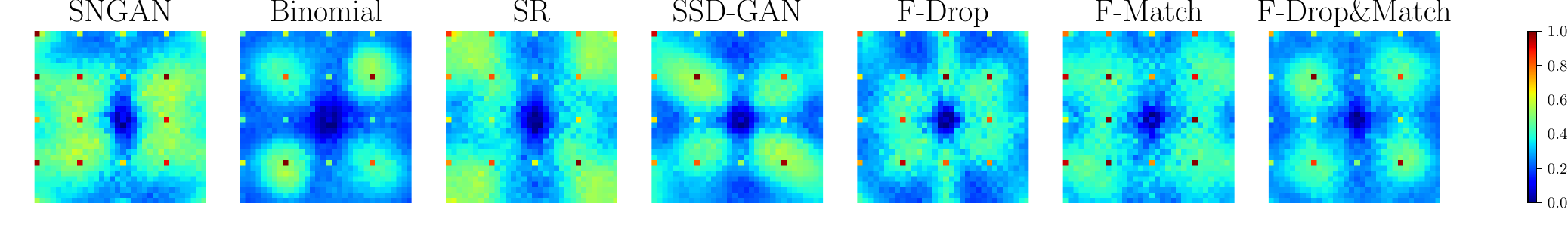}
      \end{minipage} \\
      \rotatebox[origin=c]{90}{TinyImageNet} 
      \begin{minipage}{0.95\textwidth}
         \centering
         \includegraphics[width=1.0\columnwidth]{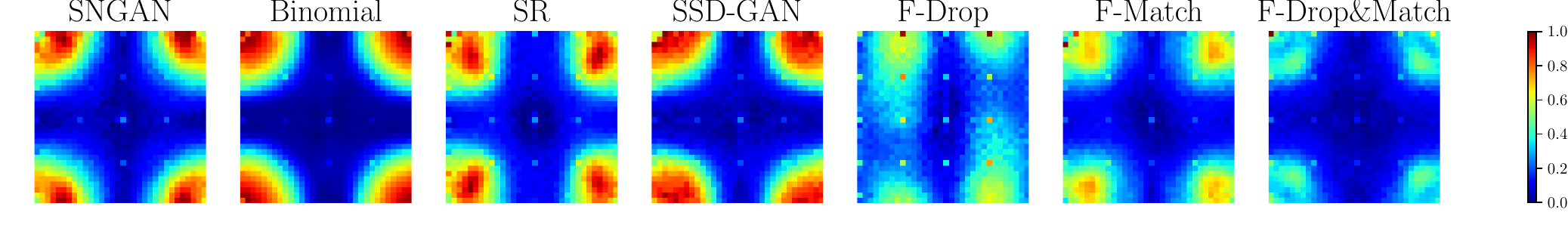}
      \end{minipage} \\
      \rotatebox[origin=c]{90}{STL-10} 
      \begin{minipage}{0.95\textwidth}
         \centering
         \includegraphics[width=1.0\columnwidth]{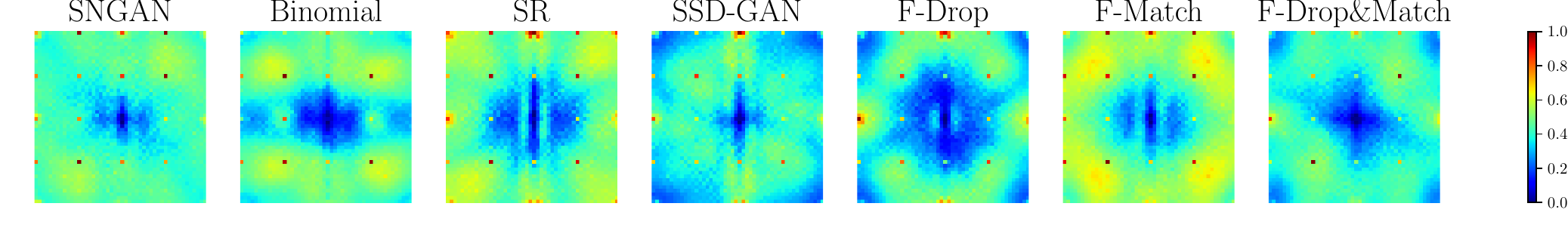}
      \end{minipage} \\
      \rotatebox[origin=c]{90}{ImageNet} 
      \begin{minipage}{0.95\textwidth}
         \centering
         \includegraphics[width=1.0\columnwidth]{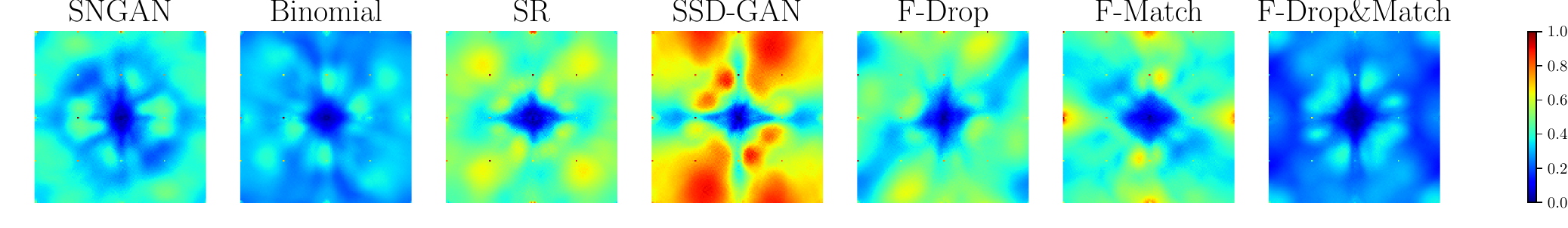}
      \end{minipage} \\
   \end{tabular}
   \caption{
         Sensitivity analysis by SFA~\cite{tsuzuku_CVPR2019_structural_sensitivity} on multiple datasets
         }
   \label{fig:sfa_multi}
\end{figure*}

In Fig.~\ref{fig:sfa_multi}, we provide the additional results of single Fourier attack (SFA) except for the results shown in Sec.~6.3 of the main paper.
We used the same visualization protocols as Sec.~6.3 of the main paper.
In all cases, our F-Drop\&Match succeeded to suppress the sensitivity to high-frequency perturbations as well as the main paper.
From the results, we consider that combining F-Drop and F-Match is quite important for the discriminators to be robust in the frequency domain.

\section{Additional Qualitative Results}\label{sec:add_qual}
\begin{figure*}
   \centering
   \begin{tabular}{lccc}
      & Real & SNGAN & F-Drop\&Match\\
      \rotatebox[origin=c]{90}{CIFAR-10} &
      \begin{minipage}{0.29\textwidth}
         \centering
         \includegraphics[width=1.0\columnwidth]{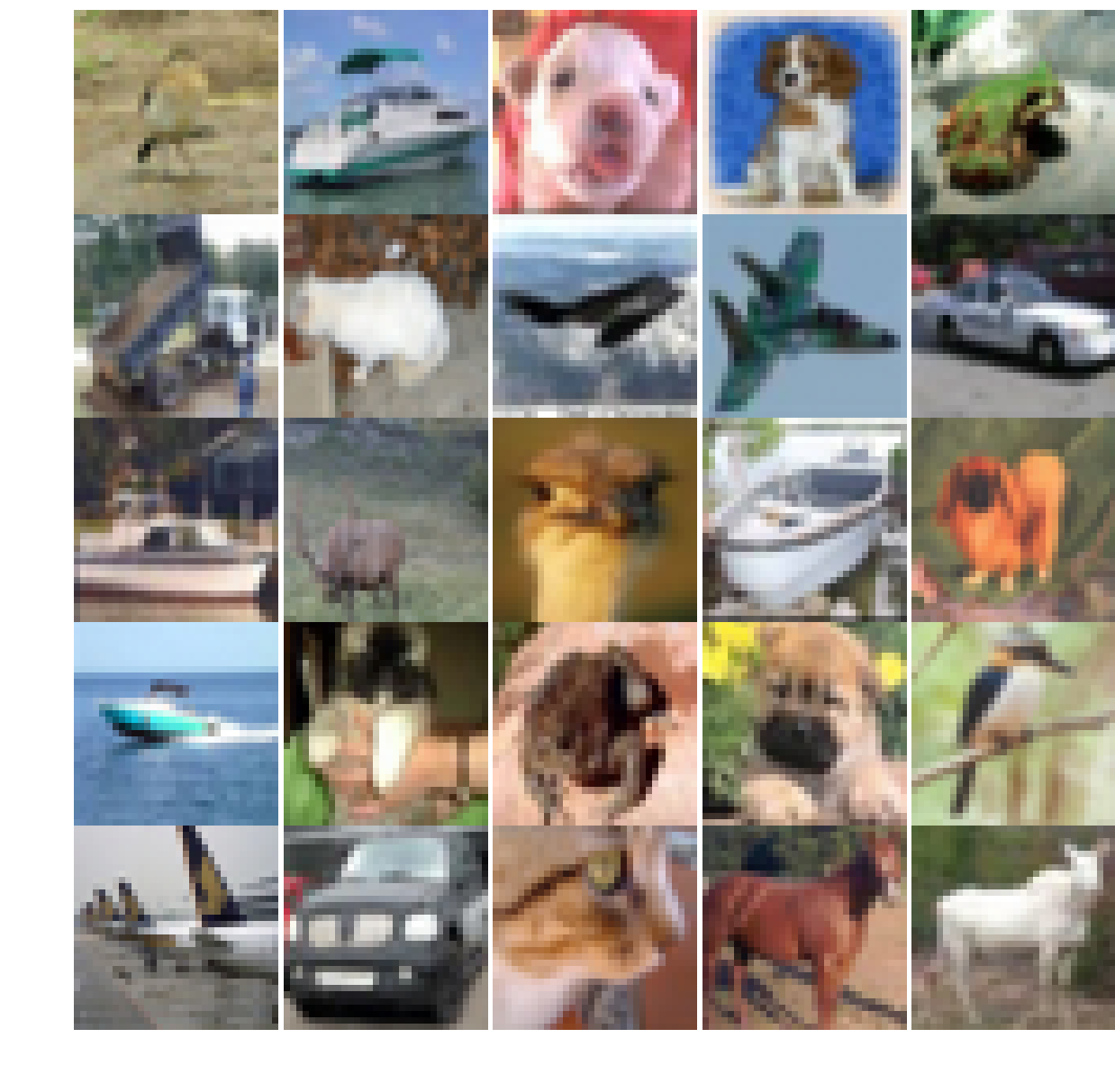} 
      \end{minipage} &
      \begin{minipage}{0.29\textwidth}
         \centering
         \includegraphics[width=1.0\columnwidth]{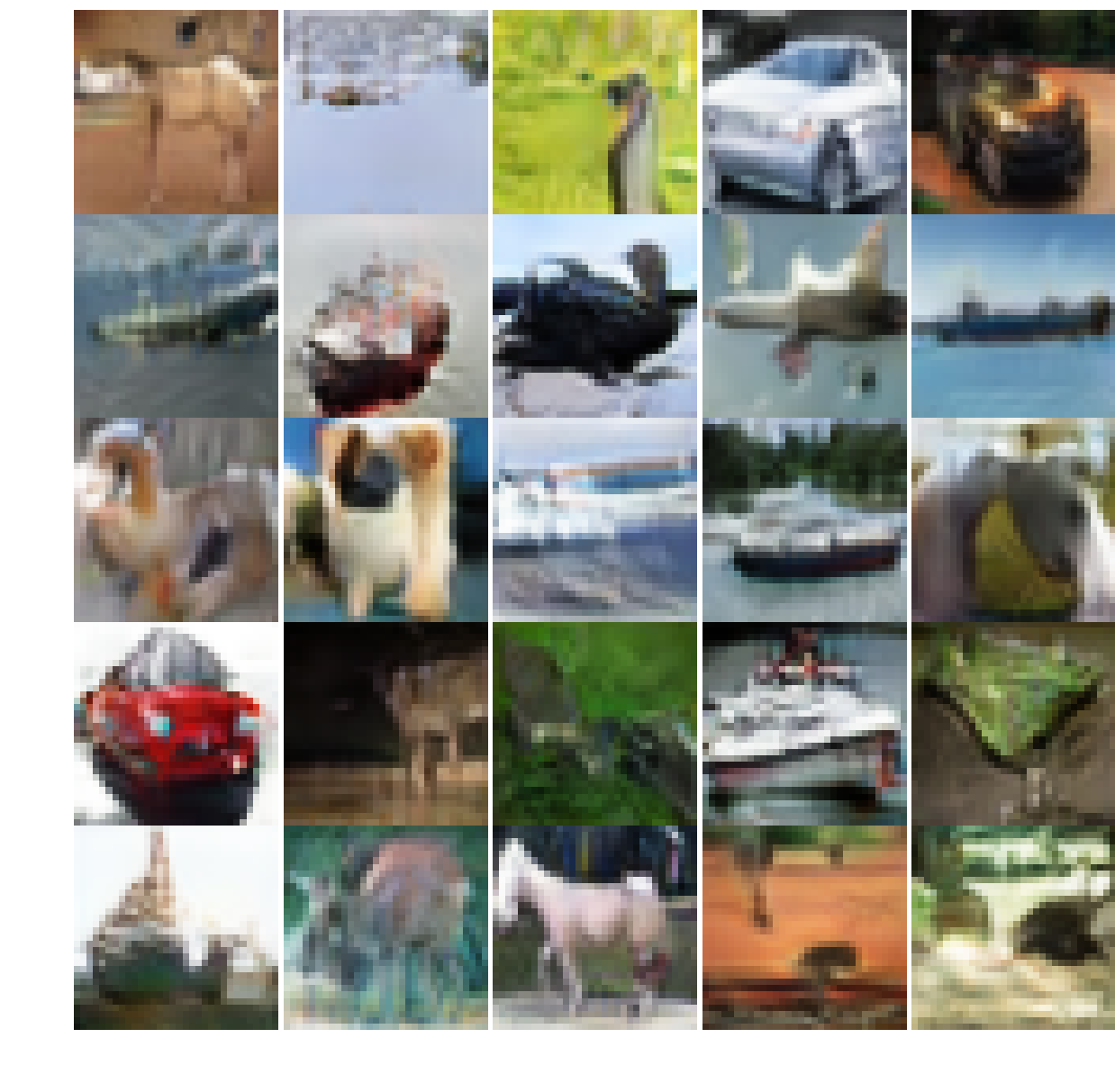} 
      \end{minipage} &
      \begin{minipage}{0.29\textwidth}
         \centering
         \includegraphics[width=1.0\columnwidth]{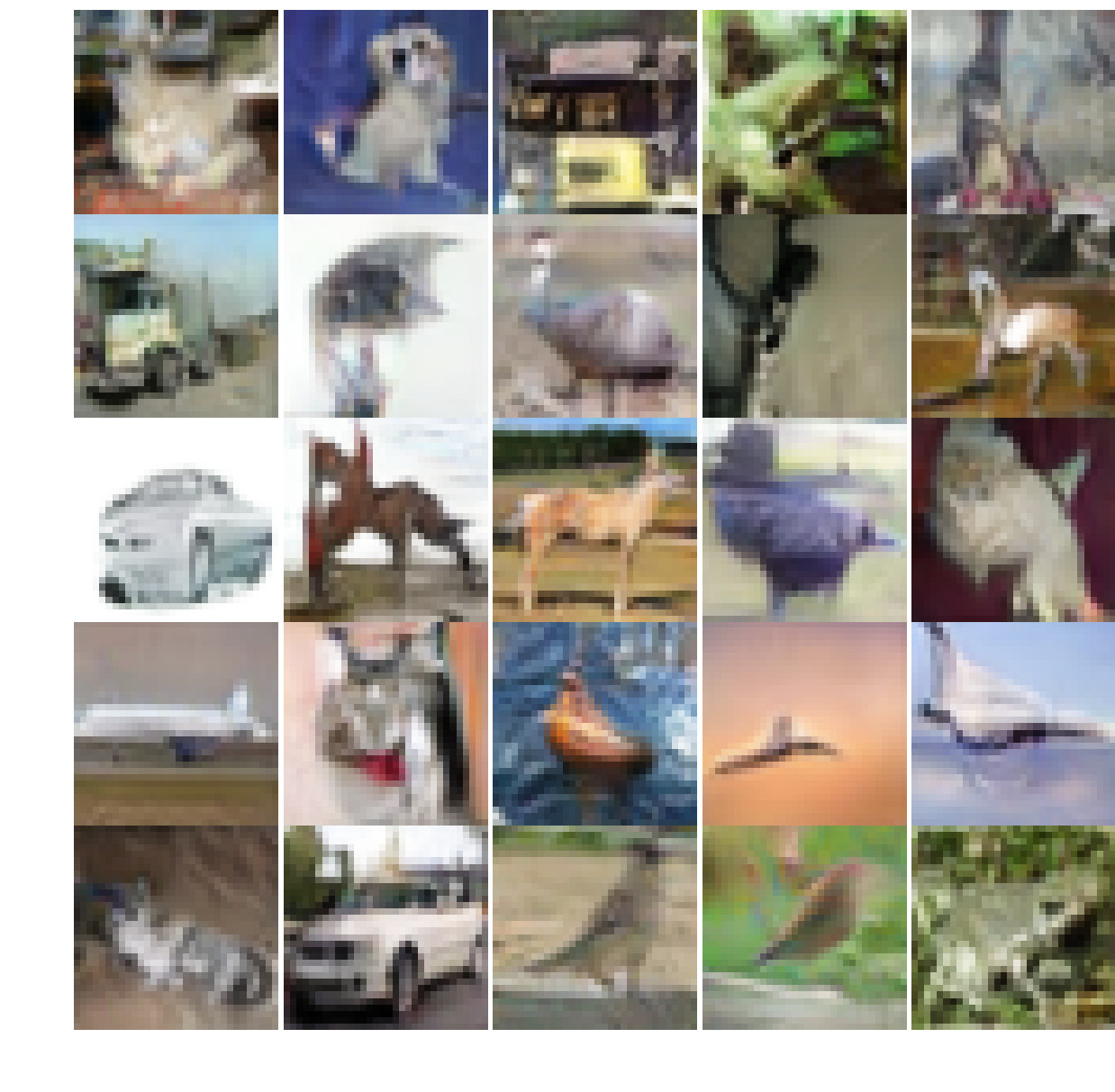} 
      \end{minipage}\\
   \end{tabular}
   \caption{
         Generated images on CIFAR-10
         }
   \label{fig:viz_cifar10}
\end{figure*}

\begin{figure*}
   \centering
   \begin{tabular}{lccc}
      & Real & SNGAN & F-Drop\&Match\\
      \rotatebox[origin=c]{90}{CIFAR-100} &
      \begin{minipage}{0.29\textwidth}
         \centering
         \includegraphics[width=1.0\columnwidth]{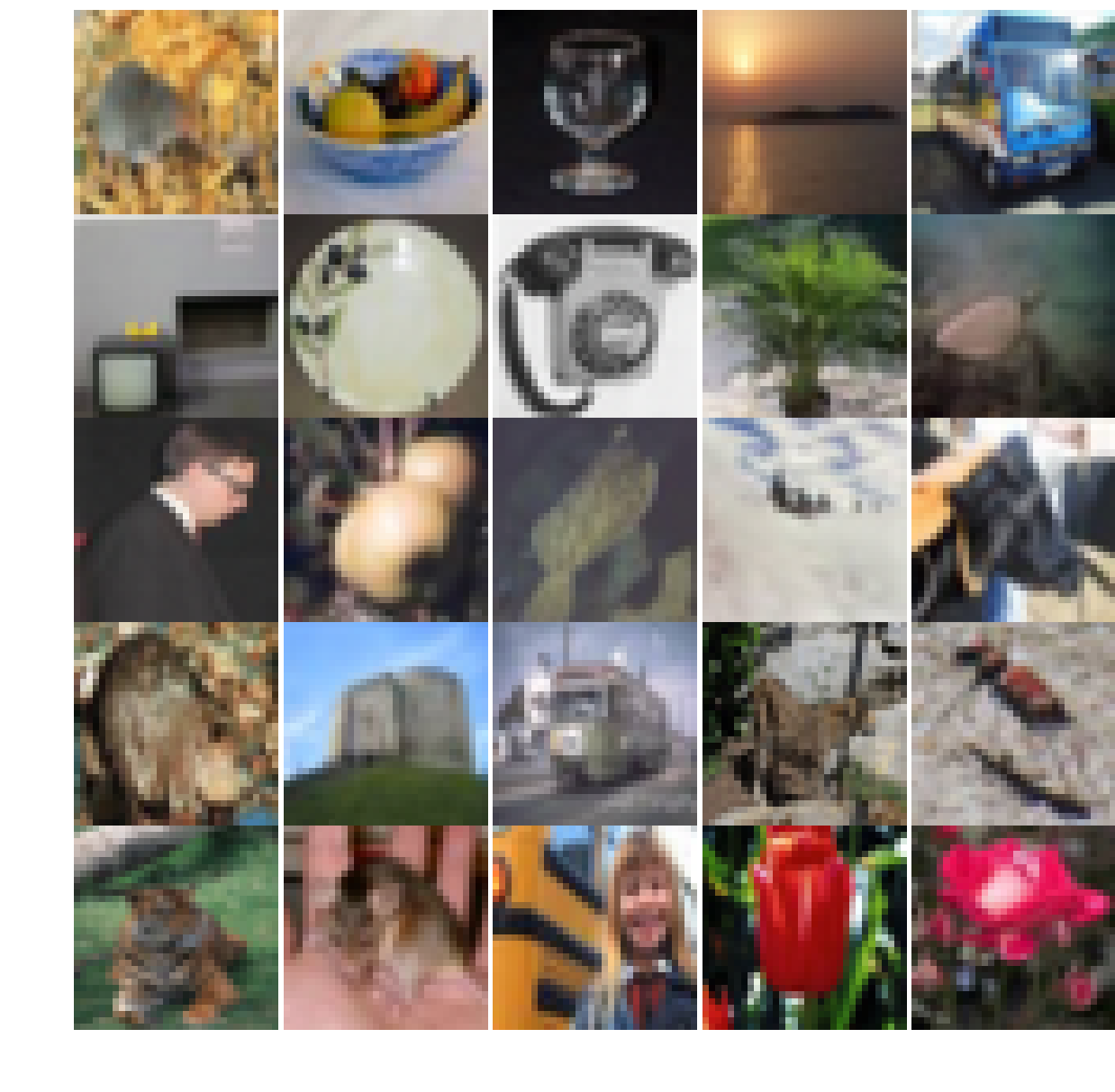} 
      \end{minipage} &
      \begin{minipage}{0.29\textwidth}
         \centering
         \includegraphics[width=1.0\columnwidth]{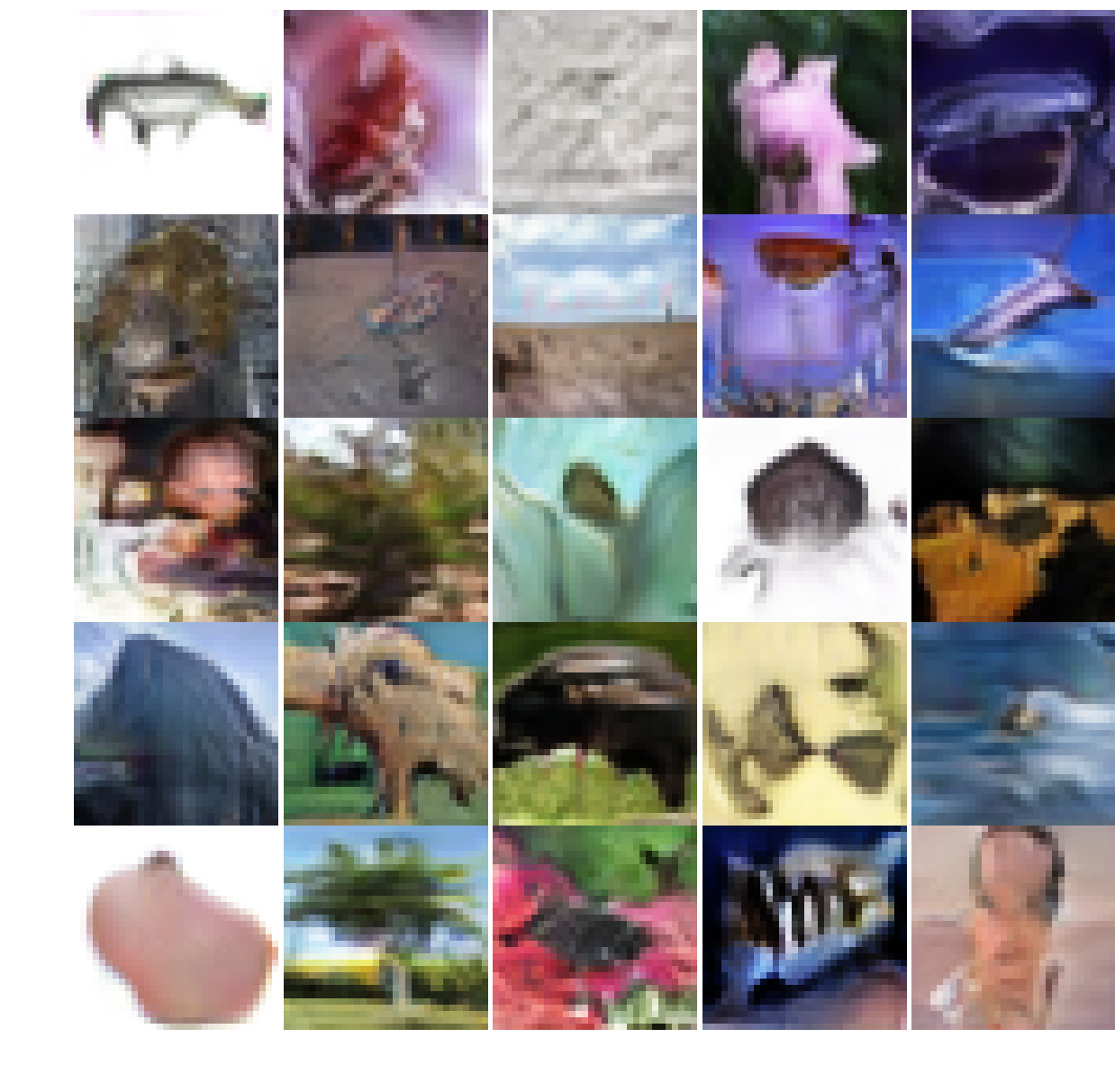} 
      \end{minipage} &
      \begin{minipage}{0.29\textwidth}
         \centering
         \includegraphics[width=1.0\columnwidth]{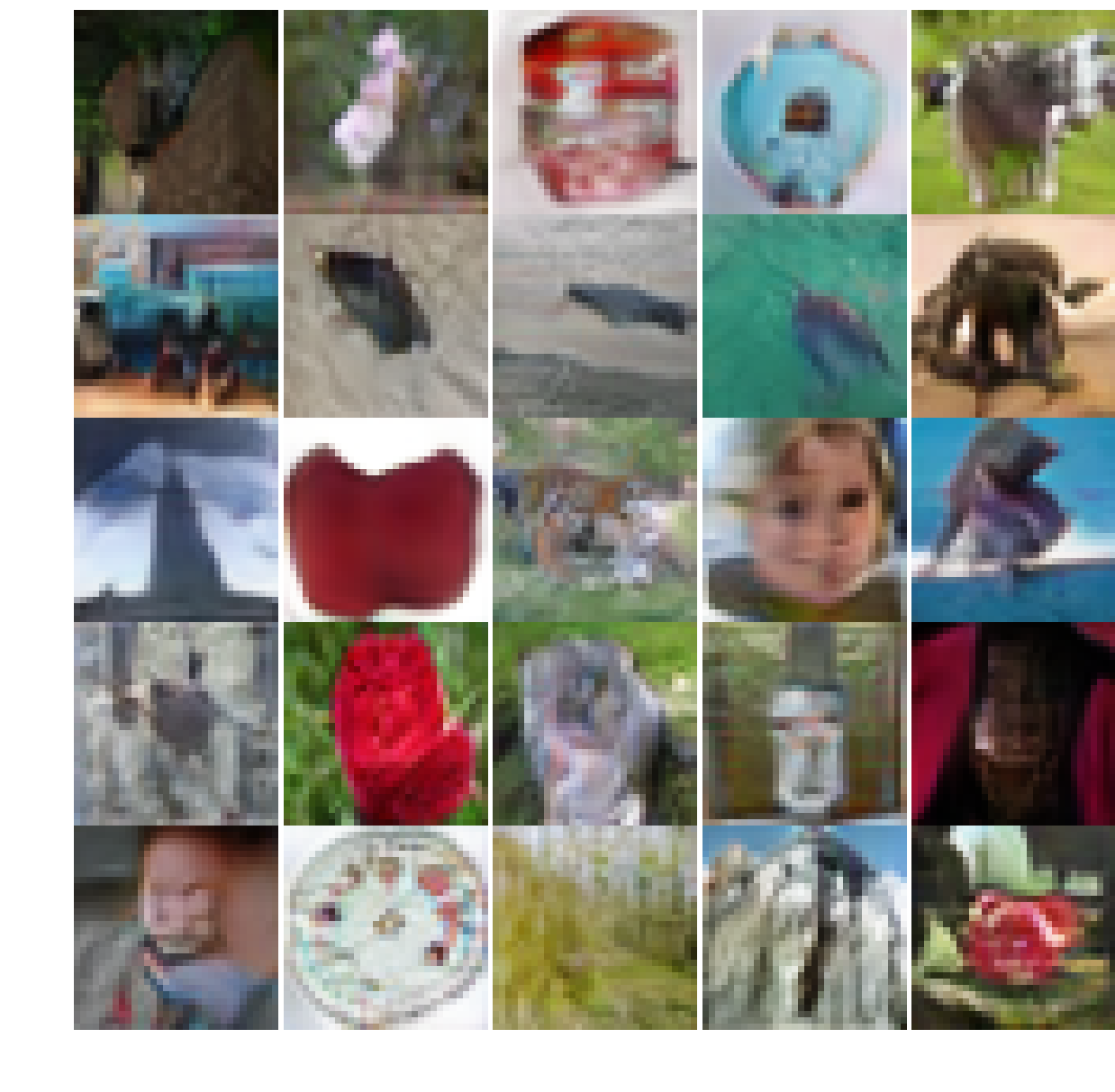} 
      \end{minipage}\\
   \end{tabular}
   \caption{
         Generated images on CIFAR-100
         }
   \label{fig:viz_cifar100}
\end{figure*}

\begin{figure*}
   \centering
   \begin{tabular}{lccc}
      & Real & SNGAN & F-Drop\&Match\\
      \rotatebox[origin=c]{90}{TinyImagenet} &
      \begin{minipage}{0.29\textwidth}
         \centering
         \includegraphics[width=1.0\columnwidth]{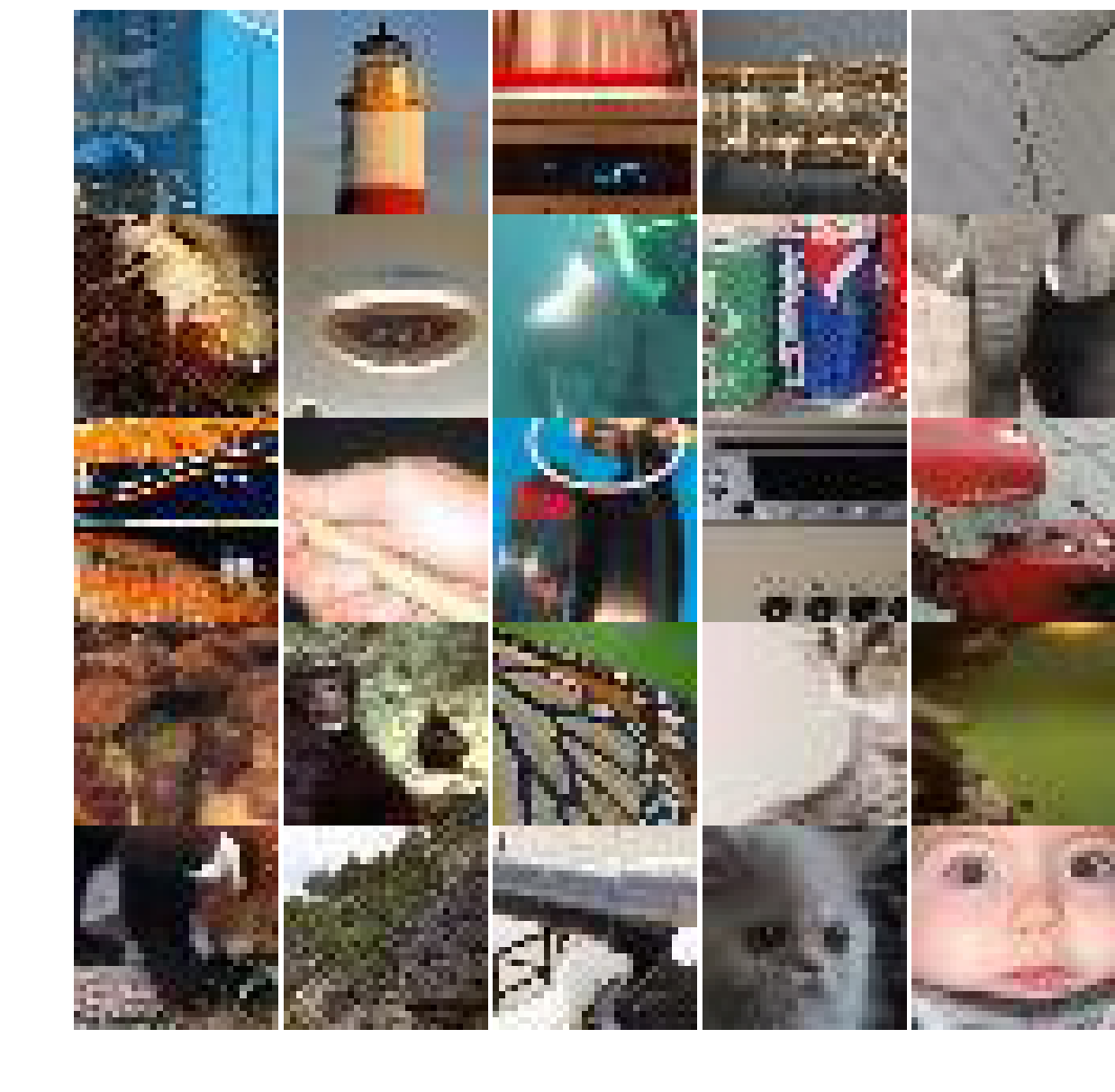} 
      \end{minipage} &
      \begin{minipage}{0.29\textwidth}
         \centering
         \includegraphics[width=1.0\columnwidth]{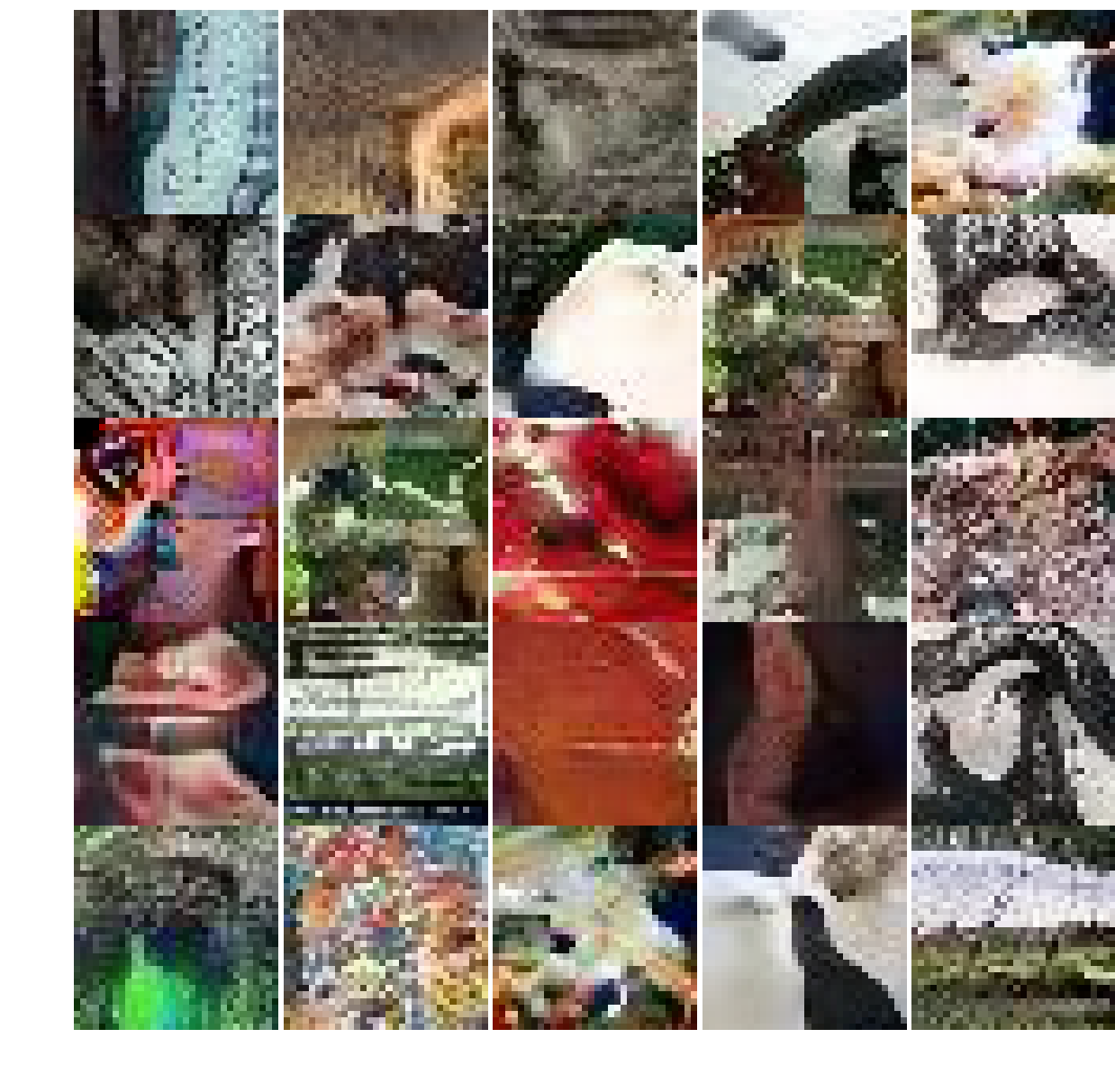} 
      \end{minipage} &
      \begin{minipage}{0.29\textwidth}
         \centering
         \includegraphics[width=1.0\columnwidth]{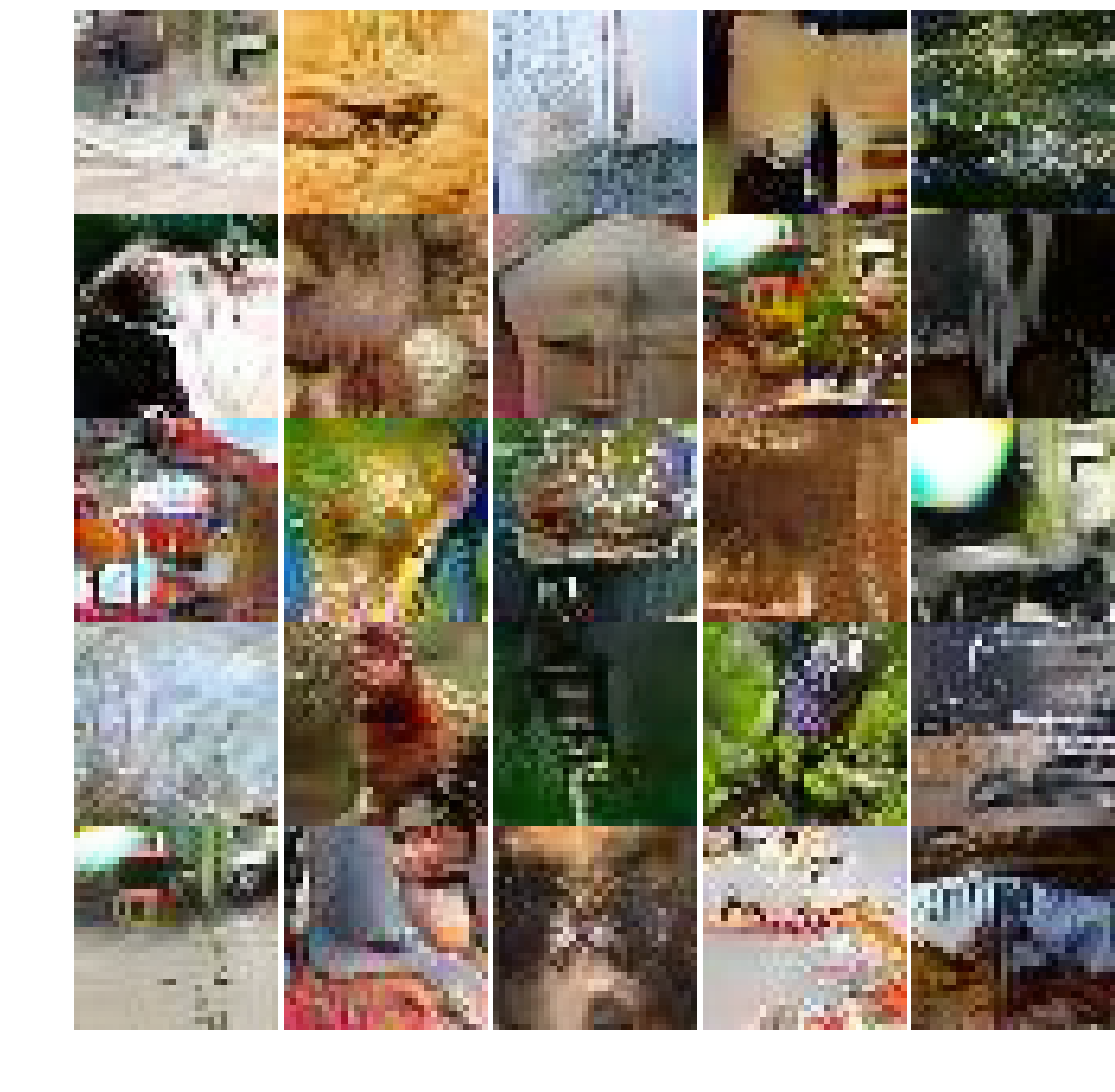} 
      \end{minipage}\\
   \end{tabular}
   \caption{
         Generated images on TinyImageNet
         }
   \label{fig:viz_tiny_imagenet}
\end{figure*}

\begin{figure*}
   \centering
   \begin{tabular}{lccc}
      & Real & SNGAN & F-Drop\&Match\\
      \rotatebox[origin=c]{90}{STL-10} &
      \begin{minipage}{0.29\textwidth}
         \centering
         \includegraphics[width=1.0\columnwidth]{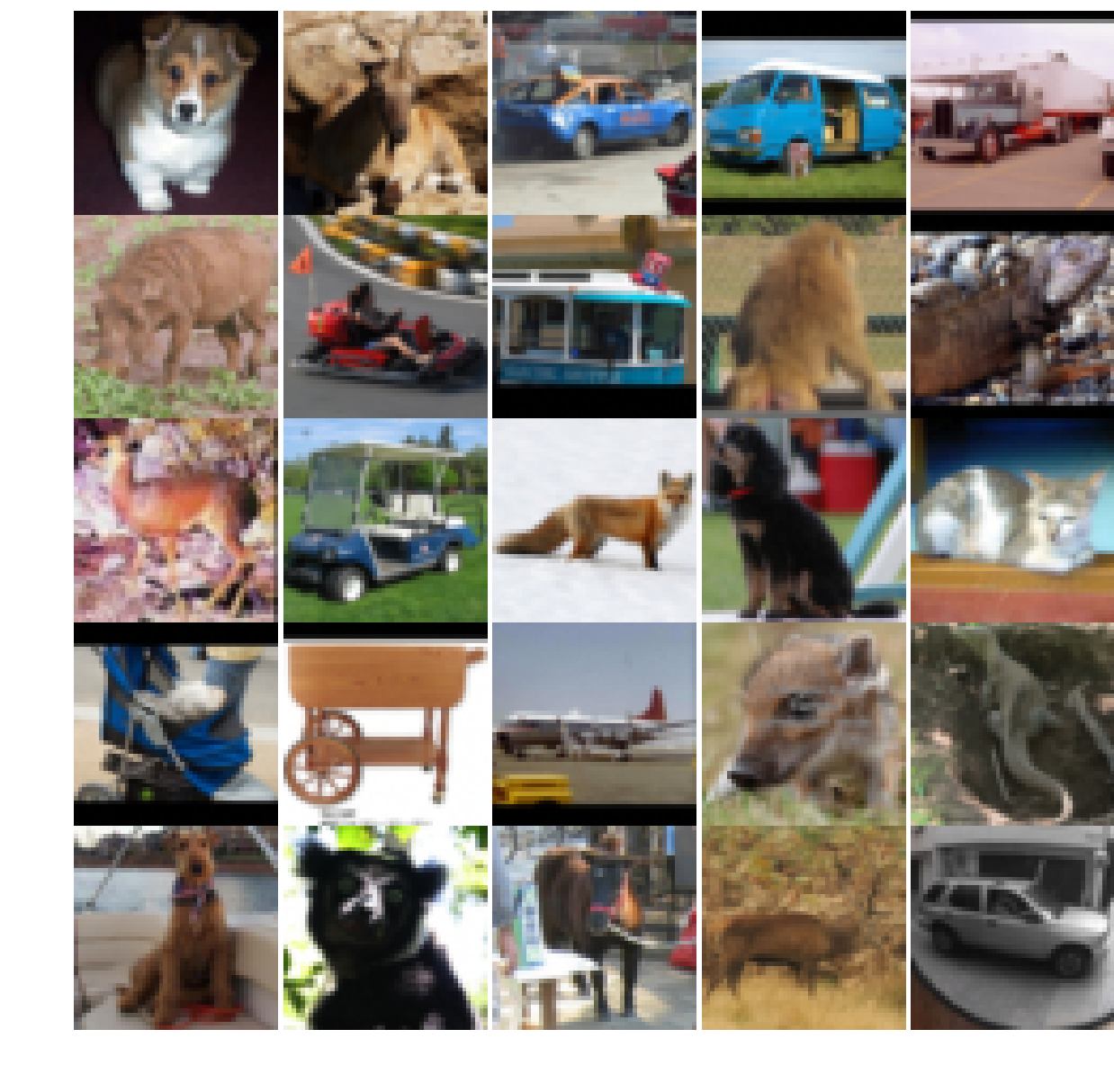} 
      \end{minipage} &
      \begin{minipage}{0.29\textwidth}
         \centering
         \includegraphics[width=1.0\columnwidth]{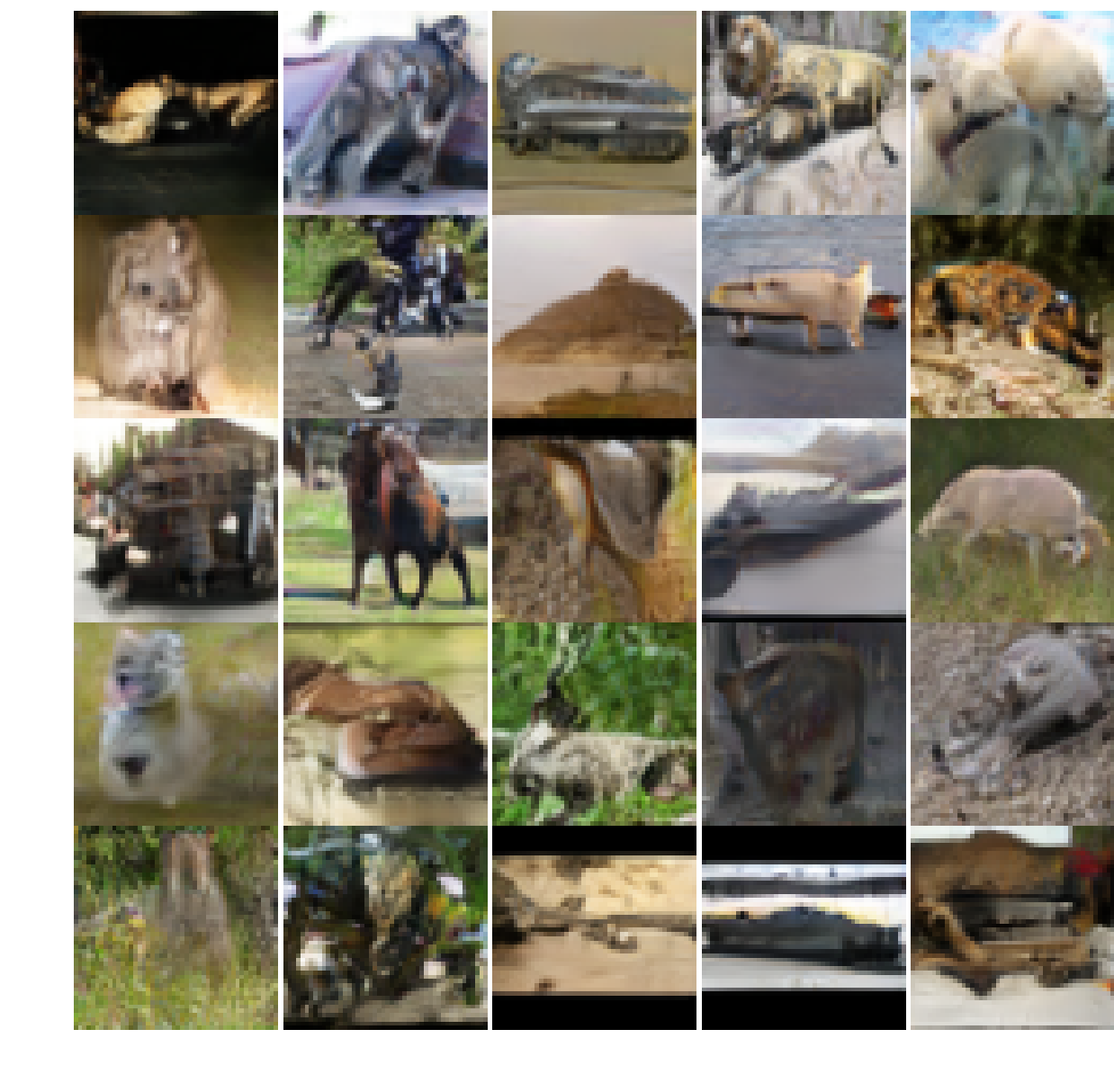} 
      \end{minipage} &
      \begin{minipage}{0.29\textwidth}
         \centering
         \includegraphics[width=1.0\columnwidth]{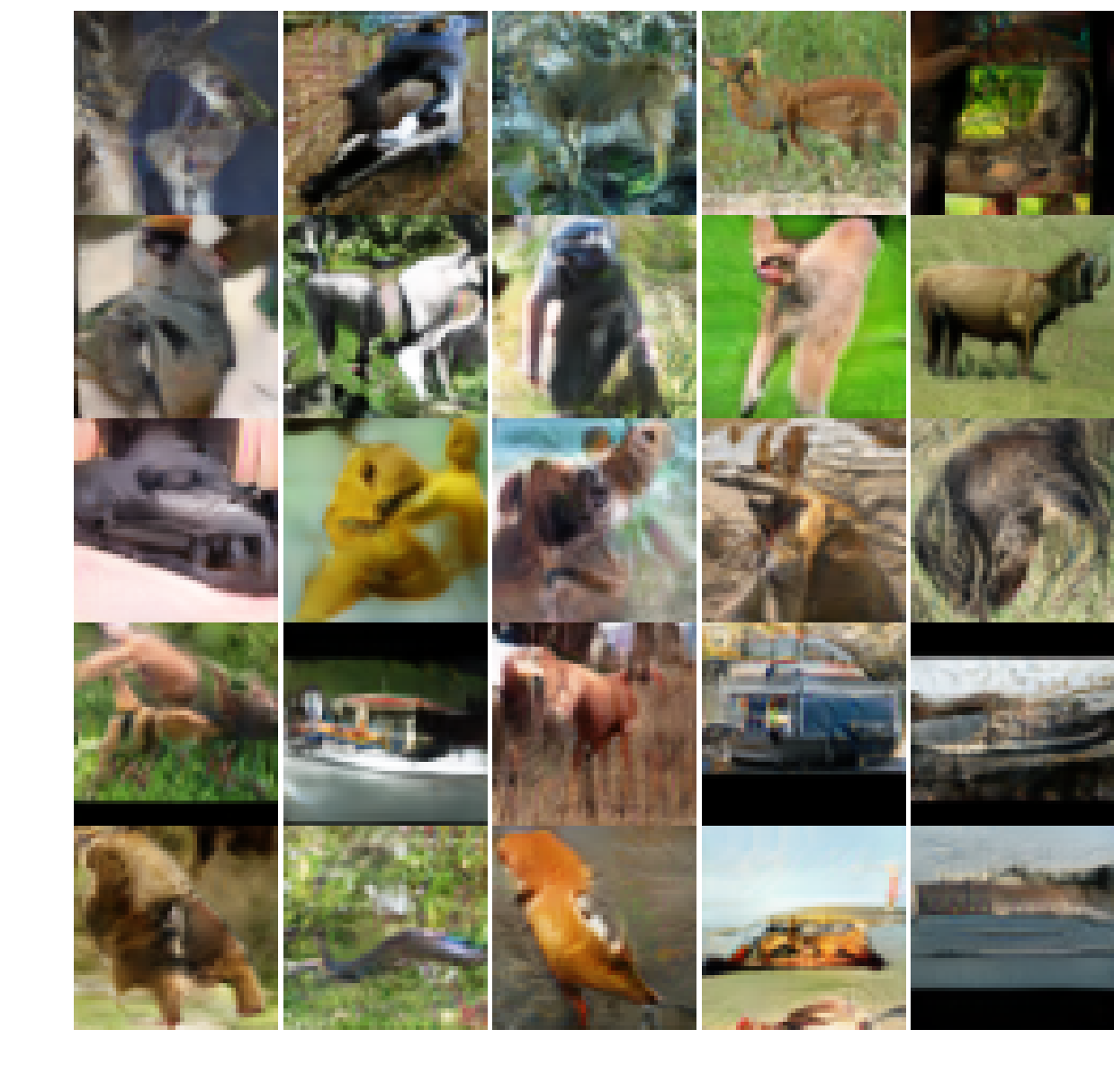} 
      \end{minipage}\\
   \end{tabular}
   \caption{
         Generated images on STL-10
         }
   \label{fig:viz_stl10}
\end{figure*}

\begin{figure*}
   \centering
   \begin{tabular}{lccc}
      & Real & SNGAN & F-Drop\&Match\\
      \rotatebox[origin=c]{90}{CelebA} &
      \begin{minipage}{0.29\textwidth}
         \centering
         \includegraphics[width=1.0\columnwidth]{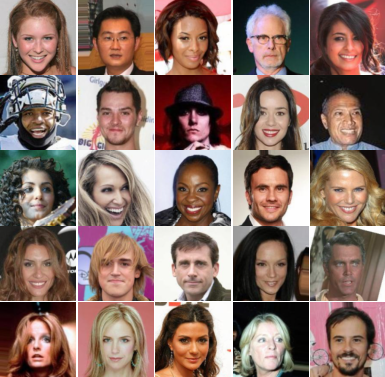} 
      \end{minipage} &
      \begin{minipage}{0.29\textwidth}
         \centering
         \includegraphics[width=1.0\columnwidth]{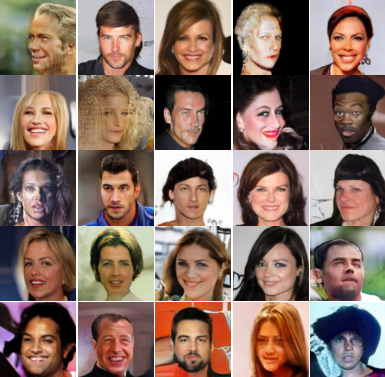} 
      \end{minipage} &
      \begin{minipage}{0.29\textwidth}
         \centering
         \includegraphics[width=1.0\columnwidth]{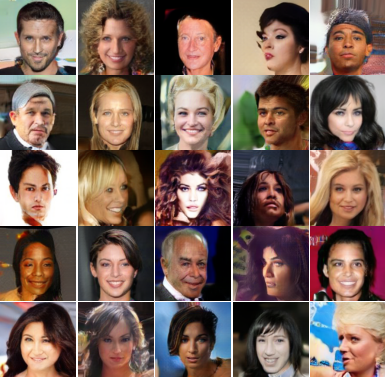} 
      \end{minipage}\\
   \end{tabular}
   \caption{
         Generated images on CelebA
         }
   \label{fig:celeba}
\end{figure*}

\begin{figure*}
   \centering
   \begin{tabular}{lccc}
      & Real & SNGAN & F-Drop\&Match\\
      \rotatebox[origin=c]{90}{ImageNet} &
      \begin{minipage}{0.29\textwidth}
         \centering
         \includegraphics[width=1.0\columnwidth]{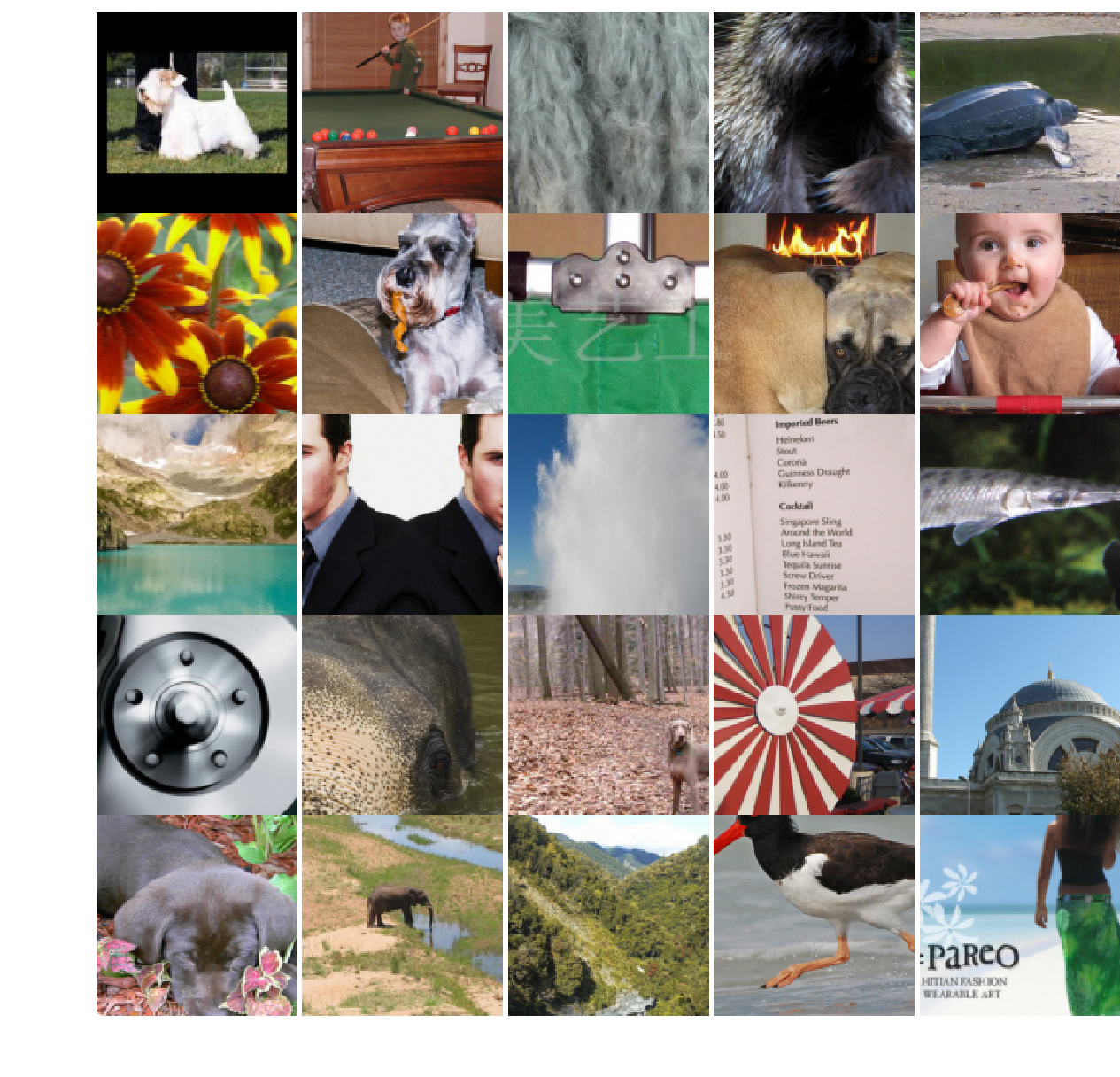} 
      \end{minipage} &
      \begin{minipage}{0.29\textwidth}
         \centering
         \includegraphics[width=1.0\columnwidth]{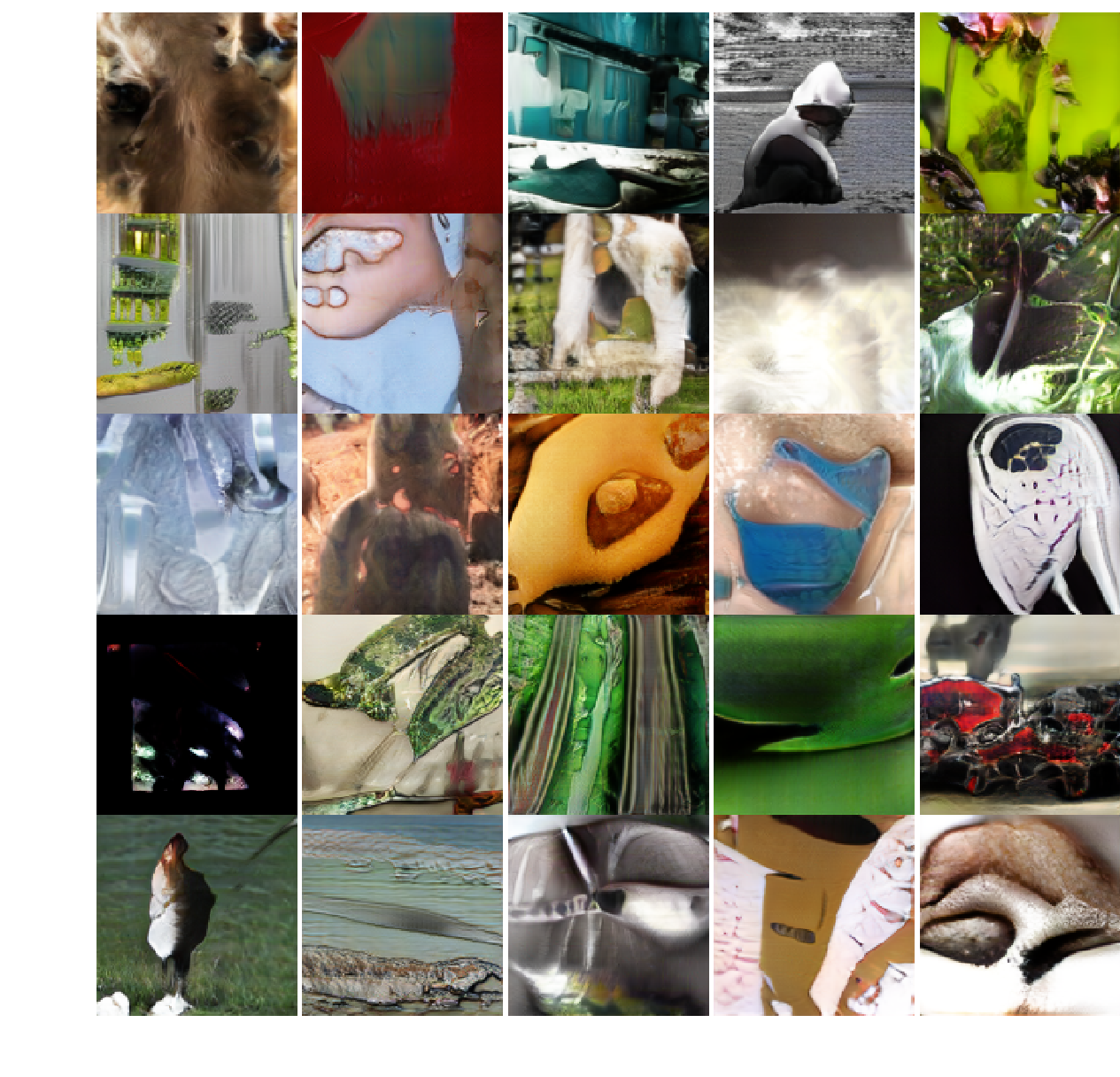} 
      \end{minipage} &
      \begin{minipage}{0.29\textwidth}
         \centering
         \includegraphics[width=1.0\columnwidth]{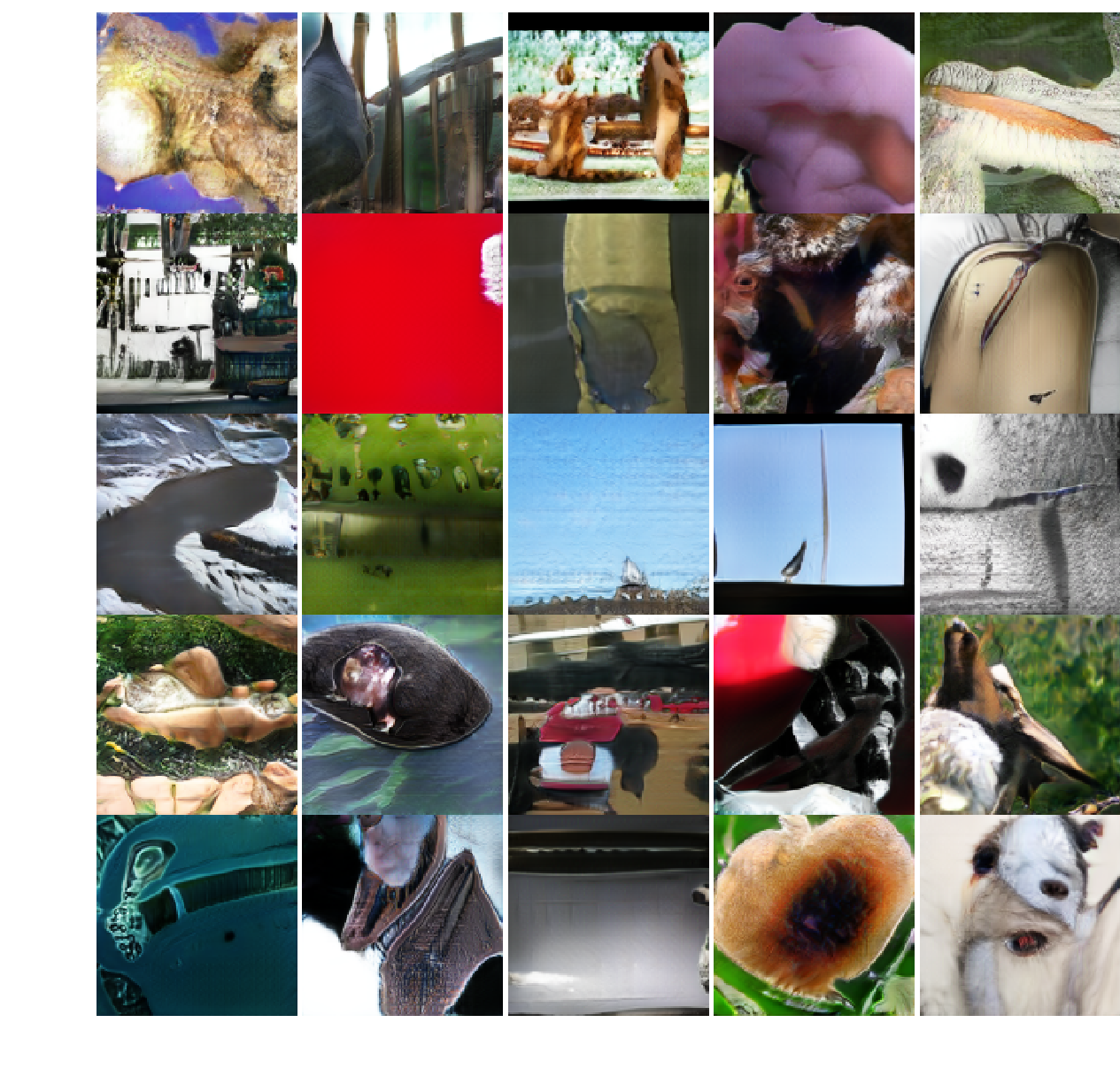} 
      \end{minipage}\\
   \end{tabular}
   \caption{
         Generated images on ImageNet
         }
   \label{fig:imagenet}
\end{figure*}

\begin{figure*}
   \centering
   \begin{tabular}{lccc}
      & Real & StyleGAN2-ADA & F-Drop\&Match\\
      \rotatebox[origin=c]{90}{FFHQ} &
      \begin{minipage}{0.29\textwidth}
         \centering
         \includegraphics[width=1.0\columnwidth]{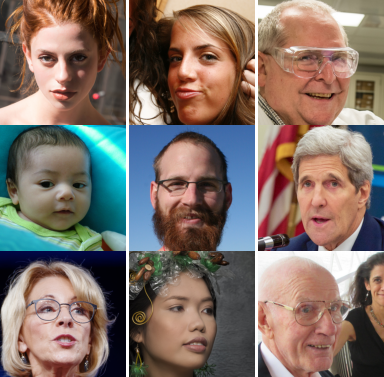} 
      \end{minipage} &
      \begin{minipage}{0.29\textwidth}
         \centering
         \includegraphics[width=1.0\columnwidth]{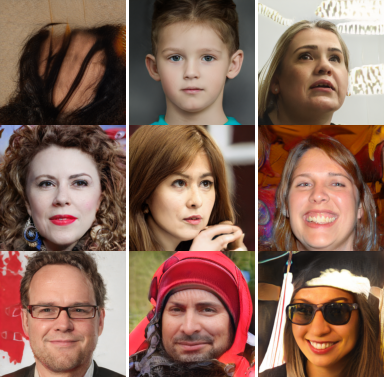} 
      \end{minipage} &
      \begin{minipage}{0.29\textwidth}
         \centering
         \includegraphics[width=1.0\columnwidth]{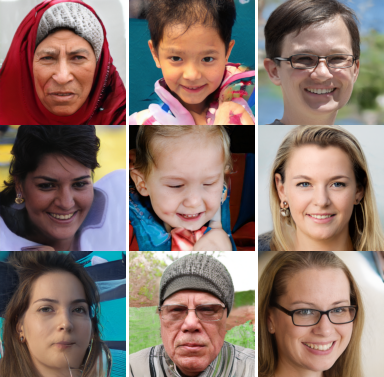} 
      \end{minipage}\\
   \end{tabular}
   \caption{
         Generated images on FFHQ (\(256\times 256\))
         }
   \label{fig:ffhq}
\end{figure*}

\begin{figure*}
   \centering
   \begin{tabular}{lccc}
      & Real & StyleGAN2-ADA & F-Drop\&Match\\
      \rotatebox[origin=c]{90}{AFHQ-Cat} &
      \begin{minipage}{0.29\textwidth}
         \centering
         \includegraphics[width=1.0\columnwidth]{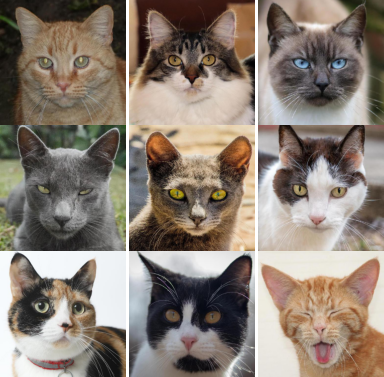} 
      \end{minipage} &
      \begin{minipage}{0.29\textwidth}
         \centering
         \includegraphics[width=1.0\columnwidth]{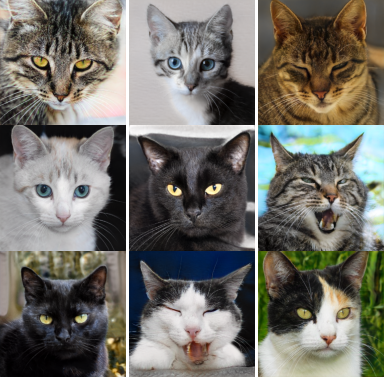} 
      \end{minipage} &
      \begin{minipage}{0.29\textwidth}
         \centering
         \includegraphics[width=1.0\columnwidth]{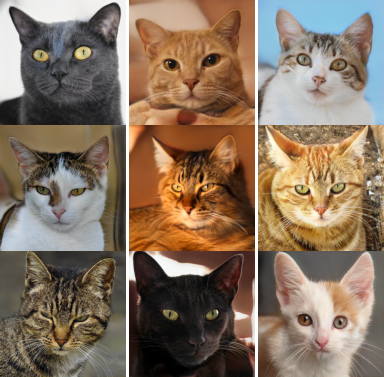} 
      \end{minipage}\\
   \end{tabular}
   \caption{
         Generated images on AFHQ-Cat (\(512\times 512\))
         }
   \label{fig:afhqcat}
\end{figure*}

\begin{figure*}
   \centering
   \begin{tabular}{lccc}
      & Real & StyleGAN2-ADA & F-Drop\&Match\\
      \rotatebox[origin=c]{90}{AFHQ-Dog} &
      \begin{minipage}{0.29\textwidth}
         \centering
         \includegraphics[width=1.0\columnwidth]{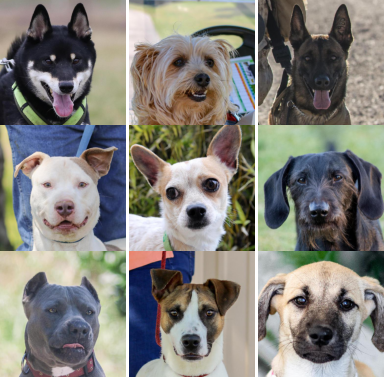} 
      \end{minipage} &
      \begin{minipage}{0.29\textwidth}
         \centering
         \includegraphics[width=1.0\columnwidth]{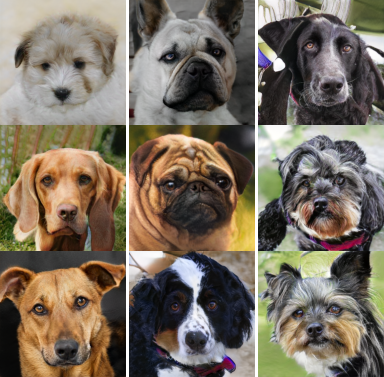} 
      \end{minipage} &
      \begin{minipage}{0.29\textwidth}
         \centering
         \includegraphics[width=1.0\columnwidth]{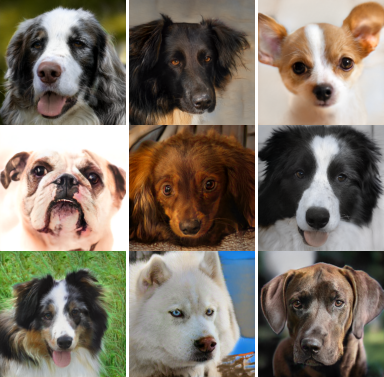} 
      \end{minipage}\\
   \end{tabular}
   \caption{
         Generated images on AFHQ-Dog (\(512\times 512\))
         }
   \label{fig:afhqdog}
\end{figure*}

\begin{figure*}
   \centering
   \begin{tabular}{lccc}
      & Real & StyleGAN2-ADA & F-Drop\&Match\\
      \rotatebox[origin=c]{90}{AFHQ-Wild} &
      \begin{minipage}{0.29\textwidth}
         \centering
         \includegraphics[width=1.0\columnwidth]{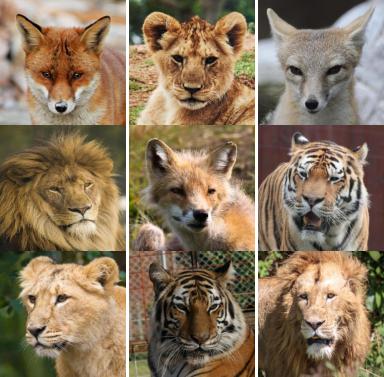} 
      \end{minipage} &
      \begin{minipage}{0.29\textwidth}
         \centering
         \includegraphics[width=1.0\columnwidth]{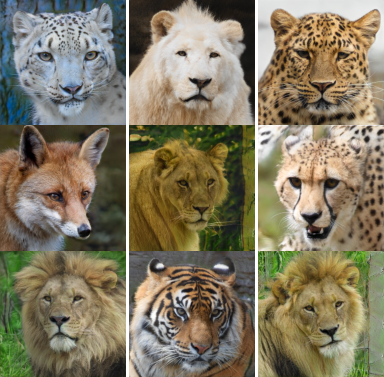} 
      \end{minipage} &
      \begin{minipage}{0.29\textwidth}
         \centering
         \includegraphics[width=1.0\columnwidth]{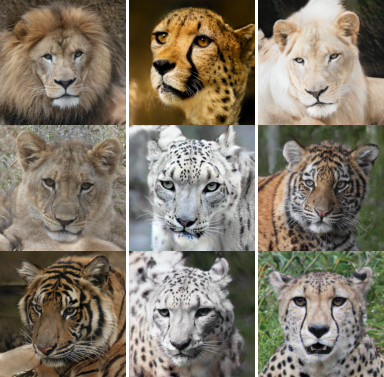} 
      \end{minipage}\\
   \end{tabular}
   \caption{
         Generated images on AFHQ-Wild (\(512\times 512\))
         }
   \label{fig:afhqwild}
\end{figure*}

We visualize the generated images from SNGAN and our F-Drop\&Match for each dataset.
Figure~\ref{fig:viz_cifar10},~\ref{fig:viz_cifar100},~\ref{fig:viz_tiny_imagenet},~\ref{fig:viz_stl10}, ~\ref{fig:celeba},~\ref{fig:imagenet},~\ref{fig:ffhq},~\ref{fig:afhqcat},~\ref{fig:afhqdog}, and~\ref{fig:afhqwild} illustrate the images.
Note that these images are randomly sampled, not cherry-picked.
As we discussed in Sec.~6.5 of the main paper, we can confirm our F-Drop\&Match succeed to synthesize detailed (high-frequency) information of images, \eg, human faces and in CIFAR-100 and textures of animal skins in STL-10.

{\small
\bibliographystyle{ieee_fullname}
\bibliography{dropmatch}
}

\end{document}